\documentclass[10pt,twocolumn,letterpaper]{article}

\usepackage{cvpr}              
\definecolor{cvprblue}{rgb}{0.21,0.49,0.74}
\usepackage[pagebackref,breaklinks,colorlinks,allcolors=cvprblue]{hyperref}
\usepackage{pifont}
\usepackage{amsmath}
\usepackage{algorithm}
\usepackage{algpseudocode}

\usepackage{booktabs,makecell,tabularx,xcolor,colortbl,adjustbox}
\usepackage{graphicx}
\usepackage{multirow}
\usepackage{array}
\usepackage{rotating}
\usepackage{diagbox}
\usepackage[table]{xcolor}

\algrenewcommand\algorithmicrequire{\textbf{Input:}}    
\algrenewcommand\algorithmicensure{\textbf{Output:}}    

\makeatletter
\algrenewcommand{\algorithmiccomment}[1]{\hfill{\footnotesize\(\triangleright\) #1}}
\makeatother

\def\paperID{10477} 
\def\confName{CVPR}
\def\confYear{2026}

\title{HyGE-Occ: Hybrid View-Transformation with 3D Gaussian and Edge Priors for 3D Panoptic Occupancy Prediction}

\author{
Jong Wook Kim\textsuperscript{\normalfont 1} \quad
Wonseok Roh\textsuperscript{\normalfont 1}\textsuperscript{$\dagger$} \quad
Ha Dam Baek\textsuperscript{\normalfont 1}\quad
Pilhyeon Lee\textsuperscript{\normalfont 3}\quad
Jonghyun Choi\textsuperscript{\normalfont 2}\quad
Sangpil Kim\textsuperscript{\normalfont 1}\thanks{Corresponding author.}\vspace{0.6em}\\
\textsuperscript{\normalfont 1}Korea University \quad
\textsuperscript{\normalfont 2}Hyundai Motor Company \quad
\textsuperscript{\normalfont 3}Inha University \quad
\vspace{-1.em}
}

\begin{document}
\maketitle
\begingroup
\renewcommand{\thefootnote}{\fnsymbol{footnote}}
\footnotetext[2]{Work done prior to joining Amazon.}
\endgroup
\begin{abstract}

3D Panoptic Occupancy Prediction aims to reconstruct a dense volumetric scene map by predicting the semantic class and instance identity of every occupied region in 3D space. Achieving such fine-grained 3D understanding requires precise geometric reasoning and spatially consistent scene representation across complex environments. However, existing approaches often struggle to maintain precise geometry and capture the precise spatial range of 3D instances critical for robust panoptic separation. To overcome these limitations, we introduce \textbf{HyGE-Occ}, a novel framework that leverages a hybrid view-transformation branch with 3D Gaussian and edge priors to enhance both geometric consistency and boundary awareness in 3D panoptic occupancy prediction. HyGE-Occ employs a hybrid view-transformation branch that fuses a continuous Gaussian-based depth representation with a discretized depth-bin formulation, producing BEV features with improved geometric consistency and structural coherence. In parallel, we extract edge maps from BEV features and use them as auxiliary information to learn edge cues.
In our extensive experiments on the Occ3D-nuScenes dataset, HyGE-Occ outperforms existing work, demonstrating superior 3D geometric reasoning.


\end{abstract}    
\section{Introduction}
\label{sec:intro}

3D Panoptic Occupancy Prediction (3DPOP) extends 2D perception tasks, such as segmentation~\cite{liu2021swin, cheng2021per, kirillov2023segment} and depth estimation~\cite{yang2024depth, yang2024depthv2, cs2018depthnet}, into a unified volumetric reasoning framework that predicts both semantic class and instance identity for every occupied region in 3D space. Unlike 2D perception~\cite{he2016deep}, which operates on visible pixels, 3DPOP reconstructs the complete spatial layout of the scene, reasoning about both visible and occluded structures. Such holistic scene understanding is essential for autonomous driving. Estimating the full 3D occupancy is vital for safe motion planning and long-term scene prediction.


\begin{figure}[t]
    \centering
    \includegraphics[width=\linewidth]{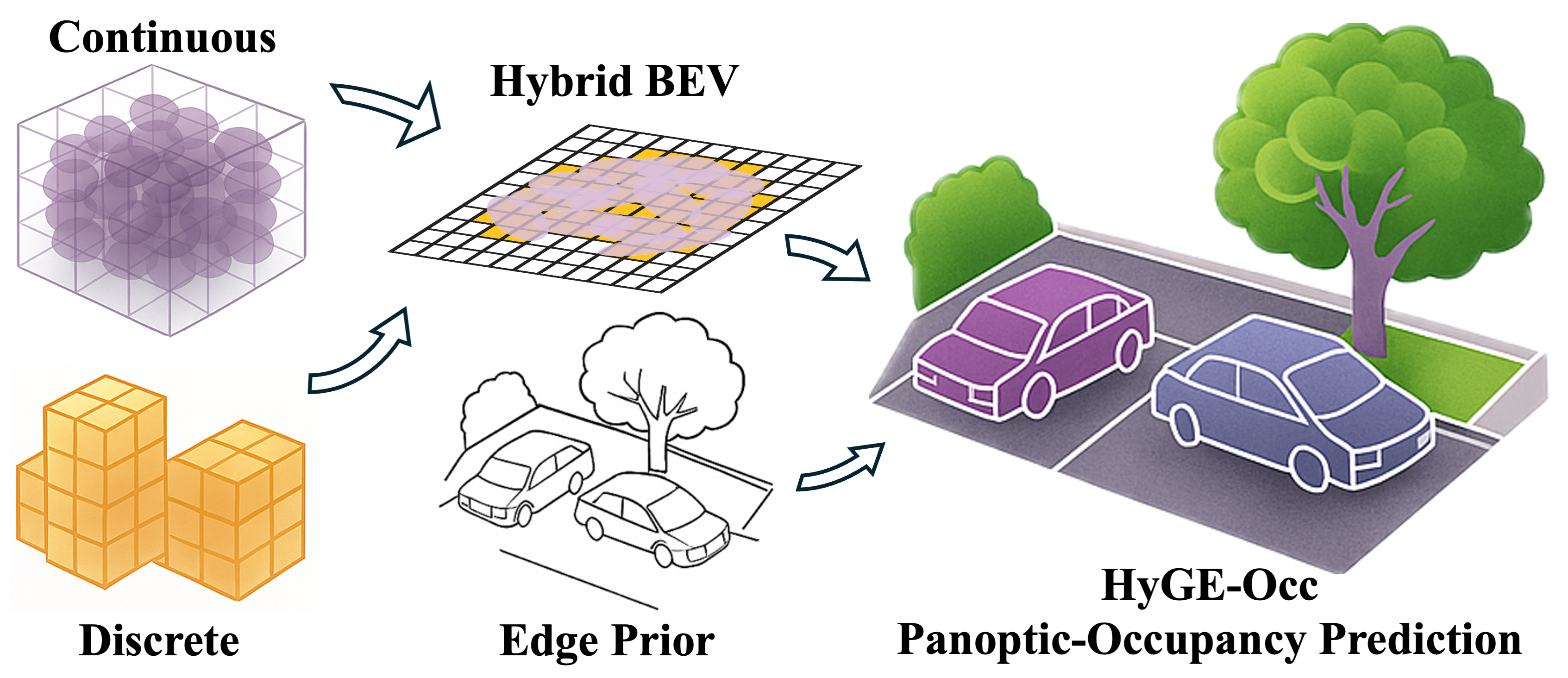}
    \vspace{-2em}
    \caption{
    HyGE-Occ fuses continuous and discrete depth representations into a hybrid BEV space and integrates an edge prior to enhance boundary cues, yielding robust 3DPOP.
    }
    \label{fig:teaser}
    \vspace{-1.5em}
\end{figure}

A key line of work in this field focuses on feature lifting from camera views to 3D space using Bird's-Eye-View~(BEV) representations, avoiding the computational burden of full 3D convolutions. The dominant paradigm, Lift-Splat-Shoot~(LSS)~\cite{philion2020lift}, performs discretized unprojection along predefined depth bins and projects image features into BEV space, providing a simple yet effective mechanism. The BEV representation captures 3D features more efficiently than heavy 3D convolutions.
Despite their advantages, discrete depth prediction methods still suffer from depth quantization artifacts, producing coarse geometry with ambiguous object edges in the 3D map~\cite{lu2025toward}.

Recent approaches introduce continuous or probabilistic representations, such as Gaussian Splatting~\cite{kerbl20233d} and triplane features~\cite{huang2023tri,cui2025loma}. These representations model 3D geometry as smooth, differentiable functions rather than discrete fixed bins, enabling them to encode structural features in a more flexible manner. In the case of Gaussian splatting, the method provides a continuous, uncertainty-aware 3D representation by predicting the mean and variance of per-pixel depth distributions, effectively characterizing both the expected depth and its surrounding ambiguity. These properties enable smooth geometric interpolation and mitigate quantization artifacts inherent to discretized depth bins. The resulting representation better captures fine-grained spatial variation and provides stable depth reasoning under occlusion or in ambiguous visual conditions, making it particularly effective for dense, complex scenes.

Motivated by this, we explore how continuous 3D Gaussian-based formulations and discretized depth reasoning can complement each other. While discretized approaches like LSS preserve precise spatial localization, continuous Gaussian representations enhance geometric smoothness and robustness to depth uncertainty. We leverage the complementary strengths of both through a unified hybrid design that balances structural precision and geometric consistency, forming a foundation of our proposed framework for improved geometric reasoning.

However, improved geometric reasoning alone does not guarantee accurate semantic or instance-level occupancy predictions. We observe that many 3D occupancy prediction frameworks~\cite{yu2024panoptic,liu2024fully} struggle to delineate semantic and instance boundaries with precision. These models misrepresent the spatial extent of objects and ambiguously localize their boundaries, leading to mislabeled edge regions or incomplete instance separation. Such errors stem from the lack of explicit supervision of inter-object boundaries, which causes boundaries that appear diffuse or uncertain.  

To address these challenges, we propose \textit{HyGE-Occ}, a hybrid view-transformation method with 3D Gaussian and Edge Priors that enhances both geometric consistency and boundary precision in 3D Panoptic occupancy prediction. HyGE-Occ introduces a \textbf{Hybrid View-Transformation Branch} that integrates dense Gaussian splatting priors with discretized LSS features through alpha blending, combining the geometric consistency of continuous representations with the local spatial precision of discretized depth reasoning. In addition, we introduce an \textbf{Edge Prediction Module} that explicitly enhances boundary awareness within BEV feature representations. Edge maps are generated from ground-truth semantic labels and serve as auxiliary supervision to strengthen boundary cues in the BEV space. This process sharpens structural details before volumetric decoding, enabling the network to produce more accurate and consistent boundaries for both semantic and instance-level occupancy predictions. Both modules are architecturally decoupled, allowing seamless integration into diverse BEV-based occupancy models without architectural redesign.

We validate HyGE-Occ on the Occ3D-nuScenes~\cite{tian2023occ3d} dataset, where it achieves state-of-the-art performance, demonstrating superior geometric fidelity and sharper panoptic delineation. When we integrate our two modules into existing frameworks such as Panoptic-FlashOcc~\cite{yu2024panoptic} and ALOcc~\cite{chen2025alocc}, we surpass their original results, confirming their effectiveness as a scalable enhancement for both panoptic and semantic occupancy prediction models.

\noindent In summary, our contributions are threefold:
\begin{itemize}
\item We propose a Hybrid View-Transformation Branch that fuses continuous and discrete BEV features for improved geometric fidelity.
\item We design an Edge Prediction Module that introduces explicit edge supervision into volumetric decoding, leading to sharper, more precise panoptic delineation.
\item We extensively evaluate our method on the Occ3D-nuScenes dataset, where quantitative and qualitative results demonstrate state-of-the-art performance, confirming the effectiveness of our approach.
\end{itemize}
\section{Related works}
\label{sec:related_works}
\textbf{3D Occupancy Prediction.}~~Early studies in 3D occupancy prediction primarily focus on estimating voxel-wise occupancy states by discretizing the scene into regular 3D grids. 
Occ3D~\cite{tian2023occ3d} established a unified benchmark across Waymo~\cite{sun2020scalability} and nuScenes~\cite{caesar2020nuscenes}, enabling consistent evaluation across sensor modalities and model architectures.
Building upon this foundation, research has progressed from LiDAR-centric pipelines~\cite{zhang2020polarnet, ye2021drinet++, ye2023lidarmultinet, ming2024occfusion, zhang2024fusionocc} toward camera-only methods~\cite{cao2022monoscene, huang2023tri, oh20253d, hou2024fastocc, ma2024cotr} that offer greater scalability and deployment efficiency.
PanoOcc~\cite{wang2023panoocc} is the first to introduce a unified voxel-based framework for camera-only 3D panoptic occupancy prediction, integrating semantic segmentation and instance-level detection into a single end-to-end model. It learns a coarse-to-fine voxel representation via multi-view and multi-frame fusion, achieving dense, panoptic-level 3D scene understanding without LiDAR input.
SparseOcc~\cite{liu2024fully} further advanced the task by introducing ray-based evaluation metrics, RayIoU and RayPQ, which measure geometric and panoptic quality along camera rays and provide a more faithful assessment of long-range scene consistency compared to voxel-only metrics.
Following these developments, Panoptic-FlashOcc~\cite{yu2024panoptic} presents a lightweight camera-only framework that operates directly on BEV features, providing an efficient and scalable 3D panoptic occupancy prediction.

\begin{figure*}[t]
    \centering
    \includegraphics[width=\linewidth]{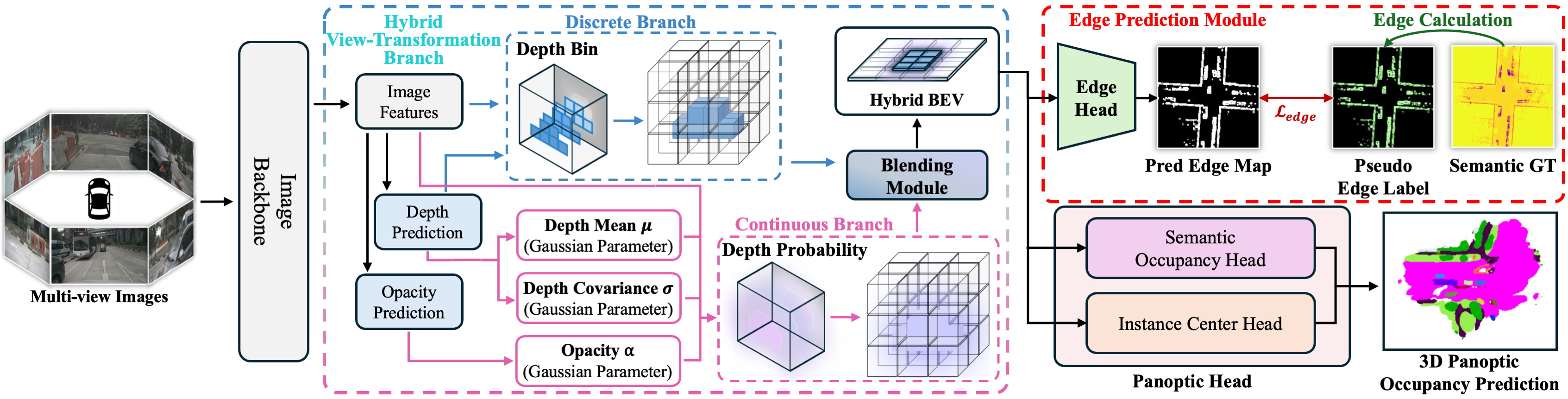}
    \vspace{-1.5em}
    \caption{
    Overview of our proposed HyGE-Occ. Our model takes multi-view images as input and first extracts image features through a shared image backbone. The \textbf{Hybrid View-Transformation Branch} fuses a discretized and continuous Gaussian-based depth representation to form hybrid BEV features that combine spatial precision and geometric continuity. The resulting BEV features are further refined by a \textbf{Edge Prediction Module}, which predicts BEV-level edge maps optimised with pseudo edge labels computed from the semantic ground truth. These boundary-enhanced BEV representations are then decoded by the panoptic head, consisting of semantic and instance center branches, to produce the final 3D panoptic occupancy prediction. 
    }
    \label{fig:main}
    \vspace{-1em}
\end{figure*}

\noindent\textbf{View Transformation for 3DOP.}~~3D occupancy prediction inherently requires transforming 2D image features into 3D space to infer volumetric scene structure.
Numerous strategies have been explored for this 2D-to-3D transformation, ranging from explicit depth-based lifting methods~\cite{philion2020lift} to implicit field modeling approaches~\cite{huang2023tri, zhang2023occformer}, and more recently, NeRF~\cite{huang2024selfocc, pan2024renderocc, zhang2025occnerf, wang2024distillnerf, boeder2025occflownet} or Gaussian Splatting~\cite{gan2025gaussianocc, jiang2025gausstr, boeder2025gaussianflowocc, zhu2025voxelsplat}-based scene representations for occupancy prediction.
Each formulation offers a distinct trade-off between geometric accuracy and spatial consistency.
Among these, Lift-Splat-Shoot (LSS)~\cite{philion2020lift} and its extensions~\cite{huang2021bevdet, li2023bevstereo, yu2023flashocc, yu2024panoptic, chen2025alocc} have been widely adopted due to their simplicity and scalability.
These methods discretize depth into uniform bins and aggregate voxel-level features to construct BEV representations.
However, such discretization inevitably introduces quantization artifacts and limits geometric continuity along the depth axis.
To overcome these limitations, GaussianLSS~\cite{lu2025toward} has proposed 3D Gaussian-based uncertainty-aware depth modeling to better handle depth ambiguity.
This suggests that the two representations can complement each other when combined, leveraging their respective strengths in spatial localization and geometric continuity.
Inspired by this insight, we propose a hybrid framework that integrates discrete with continuous depth reasoning, enhancing geometric consistency and structural coherence for 3DPOP.


\noindent\textbf{Edge Supervision.}~~
Precise boundary information provides critical cues for distinguishing semantic regions and capturing fine-grained structural details in scene understanding.
In image semantic segmentation~\cite{cheng2021boundary, lee2020structure, marin2019efficient, saeedan2021boosting}, edge-aware learning has been extensively explored to enhance boundary precision and mitigate over-smoothing near object edges.
Recently, this concept has been extended to 3D perception~\cite{gong2021boundary, roh2024edge,zhao2025bfanet}, where boundary priors have also shown effectiveness in improving geometric fidelity and spatial consistency.
These findings indicate that boundary-aware cues can enhance spatial consistency in both semantic and panoptic scene reconstruction.
Building upon this insight, we introduce an edge head that provides explicit boundary-aware guidance to the model for refined boundaries and improved panoptic consistency.

\section{Method}
\label{sec:Method}




\subsection{Model overview}
\label{sec:Model_overview}
HyGE-Occ improves BEV-based 3D panoptic occupancy prediction by introducing two novel modules, a Hybrid View-Transformation Branch and an Edge Prediction Module.
Given $N=6$ multi-view images $\{I_i\}_{i=1}^{N}$, a 2D backbone extracts feature maps $\mathbf{F}_i \in \mathbb{R}^{C \times H \times W}$. These features are then lifted into BEV space through our hybrid view-transformation process, which combines discretized $\mathbf{B}^{d}$ and continuous $\mathbf{B}^{g}$ depth features to generate hybrid BEV features $\mathbf{B}^{h}$ that better preserve geometric structure and spatial coherence. To further enhance structural precision, HyGE-Occ employs an edge prediction module trained with pseudo-edge supervision derived from semantic annotations. This module encourages clearer boundary-sensitive signals within BEV features, resulting in providing guidance to volumetric decoder towards sharper spatial delineation.
The final hybrid BEV representation is fed into a volumetric decoder to produce dense voxel-wise semantic and instance predictions. Hybrid depth fusion and boundary-aware refinement enhance the reliability and coherence of 3D panoptic occupancy predictions The model overall architecture is illustrated in Figure~\ref{fig:main}.

\subsection{Discretized Depth Unprojection}
\label{sec:LSS}
For discretized depth modeling, we divide the depth range $[d_{\min}, d_{\max}]$ into $B$ uniform intervals following the previous work~\cite{philion2020lift}, forming a discrete depth set $D$ as: 
\begin{equation}
    D = \left\{ d_i = d_{\min} + i \cdot \frac{d_{\max} - d_{\min}}{B} \right\}_{i=0}^{B-1}
\end{equation}
Each image pixel gets a predicted feature vector $\mathbf{c} \in \mathbb{R}^{C}$ and a normalized depth probability distribution 
$\boldsymbol{\alpha} \in \Delta^{B}$, where $\Delta^{B}$ is a $B$-dimensional probability simplex. For a specific depth bin $d_i$, the frustum feature 
$\mathbf{c}_{d_i} \in \mathbb{R}^{C}$ is defined by weighting the context feature by its probability: $\mathbf{c}_{d_i} = \alpha_i \, \mathbf{c}$.
These depth-weighted features are then unprojected into 3D frustum space and later aggregated into BEV coordinates.
The discretized depth branch supervises voxel occupancy through a binary cross-entropy loss as follows:
\begin{equation}
    \mathcal{L}_{\text{LSS}} = 
    -\, \mathbf{O}_{gt} \log(\mathbf{O}_{\text{LSS}}) - (1 - \mathbf{O}_{gt}) \log(1 - \mathbf{O}_{\text{LSS}}),
\end{equation}
where $\mathbf{O}_{gt}$ and $\mathbf{O}_{\text{LSS}}$ denote the ground-truth and predicted occupancy volumes, respectively.
Although the discrete unprojection mechanism efficiently lifts 2D features into 3D, it introduces unstable depth distribution and incomplete spatial coverage~\cite{lu2025toward} due to its fixed bin resolution.  
In addition, the softmax-based depth probabilities may become unstable, leading to inconsistent geometry in neighboring depths.

\begin{figure}[t]
    \centering
    \includegraphics[width=\linewidth]{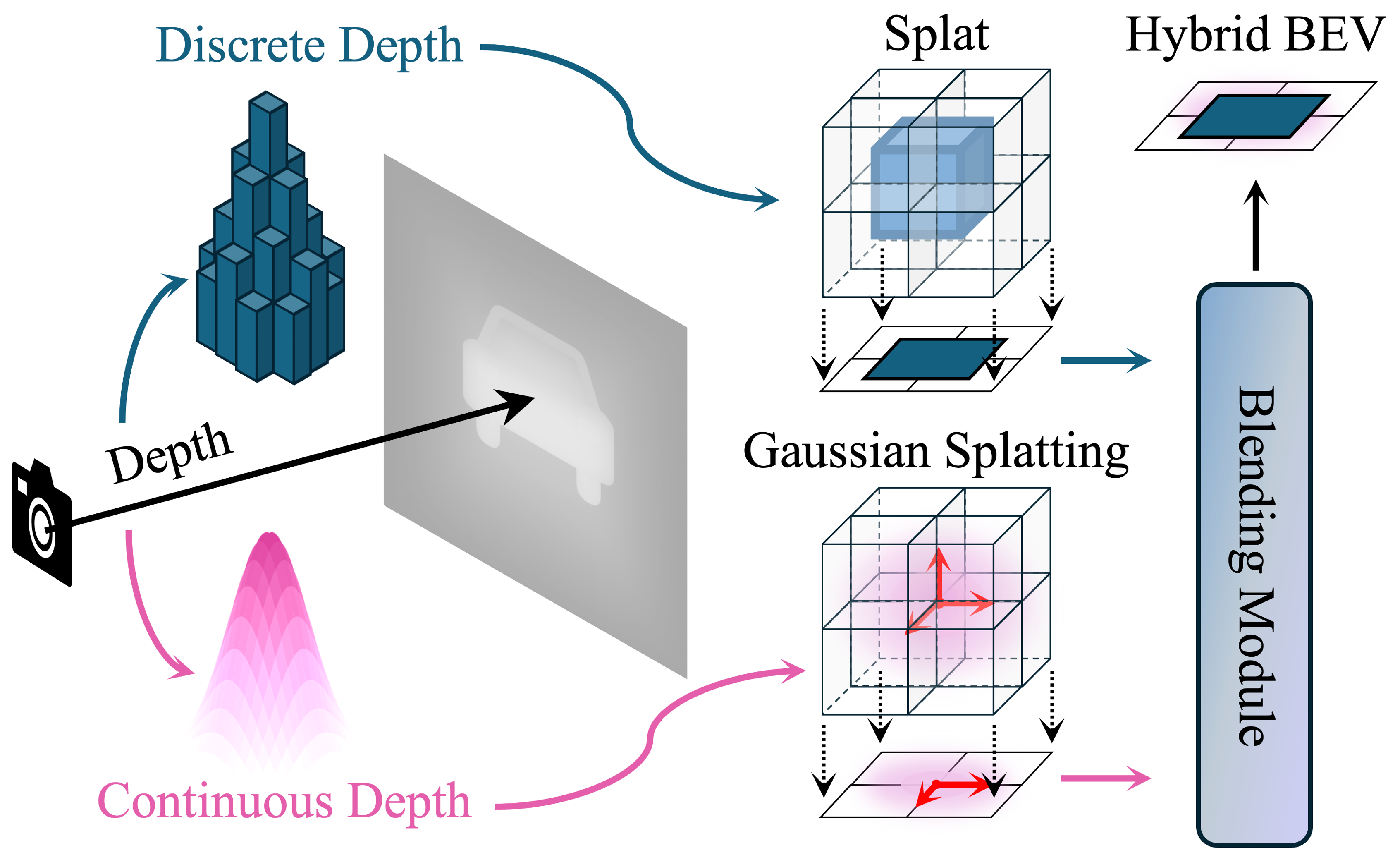}
    \vspace{-2.em}
    \caption{
   Hybrid View-Transformation Branch. The proposed module integrates both discrete and continuous depth representation through a blending module to form hybrid BEV features that combine geometric continuity with spatial precision.
    }
    \label{fig:hybrid}
    \vspace{-1em}
\end{figure}

\subsection{Continuous Gaussian-based Unprojection}
\label{sec:GSLSS}

To overcome the limitations of discrete binning, we adopt a continuous depth unprojection branch inspired by the 3D Gaussian primitives~\cite{kerbl20233d,lu2025toward}. Instead of representing depth as a hard depth bin assignment, each pixel predicts a continuous depth distribution parameterized by a mean $\mu$ and variance $\sigma^2$. This introduces an uncertainty-aware spatial range: $(\mu - k\sigma, \, \mu + k\sigma)$, where $k$ is a tolerance factor. This formulation mitigates depth quantization by allowing depth to vary smoothly within the predicted distribution, improving robustness under ambiguous conditions such as occlusion or far viewing distances.
Our use of Gaussian primitives is motivated by two key properties observed in 3D Gaussian Splatting, smooth spatial support and uncertainty-aware depth modeling. We leverage these properties to enrich the lifted 3D features, soften the influence of incorrect depth estimates, and implicitly capture object extents within the uncertainty region.

To integrate these benefits, we construct Gaussian features by mapping each pixel's predicted depth distribution to a 3D anisotropic Gaussian centered at $\mu$ with covariance $\Sigma = \mathrm{diag}(\sigma_x^2, \sigma_y^2, \sigma_z^2)$ and opacity $\alpha\in[0,1]$. 
Each Gaussian thus defines a volumetric kernel:
\begin{equation}
G_i(x) = \alpha_i \, 
\exp\!\left(-\frac{1}{2}(x - \mu_i)^{\top} \Sigma_i^{-1} (x - \mu_i)\right),
\end{equation}
where $\alpha_i$ modulates the visibility or transparency of the Gaussian primitive.  
All Gaussians are then \textit{splat-rendered} into the BEV plane through differentiable accumulation:
\begin{equation}
F_{\text{gauss}}(x) = \sum_i w_i \cdot G_i(x),
\end{equation}
where $w_i$ denotes the learned feature weight derived from the backbone. The resulting splatted representation encodes depth as a smooth spatial field, providing a continuous geometric signal that complements the discrete LSS features within our hybrid design.

In addition to occupancy estimation, the continuous unprojection branch also predicts auxiliary centerness and offset maps. These maps guide the reconstruction of object geometry in the BEV space. The centerness map $\hat{c}$ provides spatial weighting that emphasizes object centers and suppresses noisy regions. The offset map $\hat{o}$ encodes positional shifts between projected features and their ground-truth centers $c$. These additional predictions enable the Gaussian-based representation to better capture instance geometry and improve the localization accuracy of reconstructed structures. 
The continuous depth representation method is optimized with a multi-task loss that jointly supervises segmentation, centerness, and offset prediction.  
The loss is formulated as:
\begin{equation}\label{gslss_loss}
\mathcal{L}_{\text{G}} = \lambda_1 \mathcal{L}_{\text{seg}} + \lambda_2 \mathcal{L}_{\text{center}\_g} + \lambda_3 \mathcal{L}_{\text{offset}\_g},
\end{equation}
where $\lambda_1$, $\lambda_2$, and $\lambda_3$ are weighting coefficients that balance each component.  
The segmentation loss ($\mathcal{L}_{\text{seg}}$) is a focal loss~\cite{lin2018focallossdenseobject} to handle class imbalance:
\begin{equation}
\mathcal{L}_{\text{seg}} = -\alpha_t (1 - p_t)^\gamma \log(p_t),
\end{equation}
where $p_t$ is the predicted probability for the true class, $\alpha_t$ is a balancing factor, and $\gamma$ is a focusing parameter. 

The centerness loss employs an L1 objective, and the offset regression is trained with an L2 loss,
\begin{equation}
\mathcal{L}_{\text{center}\_g} = \| \hat{c} - c \|, \qquad \mathcal{L}_{\text{offset}\_g} = \| \hat{o} - o \|^2,
\end{equation}
where $\hat{c}$ and $\hat{o}$ are predicted centerness and offset values, and $c$ and $o$ are their respective ground truths.



\subsection{Hybrid View-Transformation Branch}
\label{sec:Hybrid}

While the Gaussian formulation provides continuous, uncertainty-aware estimation across depth, the LSS branch preserves local contrast and details through explicit bin-wise formation.
To leverage the complementary properties, as shown in Figure~\ref{fig:hybrid}, we fuse BEV features from both branches, with a simple yet effective $\alpha$-blending strategy. This hybrid representation captures Gaussian continuity and LSS spatial precision, producing stable, structurally coherent BEV embeddings.
For each view $i$, we obtain the following BEV features as:
\begin{equation}
    \mathbf{B}_i^{d} = \mathcal{U}_{\text{LSS}}(\mathbf{F}_i), \qquad
    \mathbf{B}_i^{g} = \mathcal{U}_{\text{G}}(\mathbf{F}_i),
\end{equation}
where $\mathbf{B}_i^{d}$ and $\mathbf{B}_i^{g}$ denote BEV features from discretized $\mathcal{U}_{\text{LSS}}$ and continuous $\mathcal{U}_{\text{G}}$ projections, respectively.  
Then they are $\alpha$-blended into the hybrid BEV representation $\mathbf{B}_i^{h}$:
\begin{equation}
    \mathbf{B}_i^{h} = \alpha \, \mathbf{B}_i^{g} + \, \mathbf{B}_i^{d},
    \quad \alpha \in [0,1].
\end{equation}
All hybrid features from $N$ views are finally fused into $\mathbf{B}_{\text{agg}}$ via a standard BEV aggregation $\mathrm{Fuse}$:
\begin{equation}
    \mathbf{B}_{\text{agg}} = \mathrm{Fuse}\big(\{\mathbf{B}_i^{h}\}_{i=1}^{N}\big).
\end{equation}
The obtained geometry-aware BEV embedding serves as input to the volumetric decoder.


\begin{algorithm}[t]
\caption{Edge Prediction Module}
\label{alg:edge_head}
\begin{algorithmic}[1]
\Require Semantic ground truth $\mathbf{Y}_{gt} \in \mathbb{R}^{H \times W}$;
\Statex BEV feature map $\mathbf{B}_{\text{bev}} \in \mathbb{R}^{C \times H \times W}$
\Ensure Edge loss $\mathcal{L}_{\text{edge}}$
\Statex \textbf{Ground-truth edge extraction}
\State $\mathbf{E}_{gt} \gets \sqrt{(S_x * \mathbf{Y}_{gt})^2+(S_y * \mathbf{Y}_{gt})^2}$ \Comment{Pseudo Edge label}
\Statex \textbf{BEV edge prediction}
\State $\mathbf{E}_{pred} \gets \mathcal{F}(\mathbf{B}_{\text{bev}})$ \Comment{Predict edge probability map}

\Statex \textbf{Edge loss calculation}
\State $\mathcal{L}_{\text{edge}} \gets \mathrm{BCE}(\mathbf{E}_{pred}, \mathbf{E}_{gt})$
\State \Return $\mathcal{L}_{\text{edge}}$
\end{algorithmic}
\end{algorithm}

\subsection{Edge Prediction Module}
\label{sec:Edge}
Recent occupancy prediction frameworks primarily optimize volumetric or BEV losses that encourage accurate occupancy predictions, but often overlook explicit boundary reasoning. 
As a result, they tend to produce blurry or inconsistent object borders, especially where semantic classes or instance identities intersect.
To refine structural precision, we introduce the edge prediction module that explicitly encourages BEV features with structural cues and provides boundary-aware bev feature to the volumetric decoder.
The pseudo-code showing the steps of boundary supervision by the edge prediction module is presented in Algorithm~\ref{alg:edge_head}.
The edge head $\mathcal{F}$, a mlp-based projection with activation function $\sigma$, operates on intermediate BEV features to predict a BEV-level edge probability map $\mathbf{E}_{pred}$.
\begin{equation}
    \mathbf{E}_{pred} = \mathcal{F}(\mathbf{B}_{bev}),~~\mathcal{F}=W\sigma(W\mathbf{x}),
\end{equation}
which strengthens structural discontinuities in the reconstructed scene. 
Ground-truth edges $\mathbf{E}_{gt}$ are derived from panoptic annotations via Sobel filtering~\cite{kanopoulos1988design} as:
\begin{equation}
    \mathbf{E}_{gt} = \sqrt{(S_x * \mathbf{Y}_{gt})^2 + (S_y * \mathbf{Y}_{gt})^2},
\end{equation}
where $*$ denotes convolution, $S_x$ and $S_y$ are Sobel kernels, and $\mathbf{Y}_{gt}$ is the semantic ground truth. 
The edge head is optimized with a binary cross-entropy loss as follows:
\begin{equation}
    \mathcal{L}_{\text{edge}} = \mathrm{BCE}(\mathbf{E}_{pred}, \mathbf{E}_{gt}),
\end{equation}
providing auxiliary boundary supervision that guides the volumetric decoder toward sharper contours between adjacent semantic regions and more consistent instance delineation.
By explicitly encouraging sharper structural boundaries, the proposed edge head improves instance separation and enhances overall panoptic delineation while introducing negligible computational overhead.

\subsection{Panoptic Occupancy Prediction Head}

Following the convention~\cite{yu2024panoptic}, HyGE-Occ predicts dense voxel-wise semantic and instance-aware occupancy volumes from the hybrid BEV feature representation.  
The prediction head consists of three branches that infer semantic occupancy, instance center heatmaps, and offset vectors, which jointly enable complete 3D panoptic reconstruction.

\noindent\textbf{Semantic Occupancy.}
The semantic branch estimates per-voxel categorical probabilities $\hat{S} \in \mathbb{R}^{H \times W \times D \times C}$, where $C$ denotes the number of semantic classes.  
This branch uses a lightweight 3D convolutional decoder that aggregates multi-scale BEV features from the hybrid Gaussian--LSS encoder.  
The semantic loss is computed as the standard cross-entropy objective:
\begin{equation}
\mathcal{L}_{\text{sem}} = - \sum_{x,y,z} \log \hat{S}_{x, y, z}^{(C_{gt})},
\end{equation}
where $c_{gt}$ is the ground-truth class label for voxel $(x, y, z)$.

\noindent\textbf{Instance Center and Offset Prediction.}
To separate instances within the same semantic category, HyGE-Occ predicts an instance center heatmap $\hat{C}$ and a per-voxel offset vector $\hat{O} \in \mathbb{R}^3$.  
The center heatmap encodes the likelihood of each voxel being close to an instance centroid, while the offset vector points from each voxel to its corresponding center.  
These values are supervised by an L1 loss:
\begin{equation}
\mathcal{L}_{\text{center}} = \| \hat{C} - C_{gt} \|, \qquad \mathcal{L}_{\text{offset}} = \| \hat{O} - O_{gt} \|,
\end{equation}
where $C_{gt}$ and $O_{gt}$ are the ground-truth center and offset fields derived from 3D instance annotations.

\begin{table*}[t]
\centering
\resizebox{\textwidth}{!}{
\begin{tabular}{l|c|c|c|c|ccc|c|c}
\toprule
\textbf{Method} & \textbf{Backbone} & \textbf{Input Size} & \textbf{Vis. Mask} & 
\textbf{RayIoU} & \textbf{RayIoU$_{1m}$} & \textbf{RayIoU$_{2m}$} & \textbf{RayIoU$_{4m}$} & 
\textbf{mIoU} & \textbf{FPS} \\
\midrule
OccFormer ~\cite{zhang2023occformer}  & ResNet-101 & 928$\times$1600  & \ding{51} & - & - & - & - & 21.9 & - \\
CTF-Occ ~\cite{tian2023occ3d}  & ResNet-101 & 928$\times$1600  & \ding{51} & - & - & - & - & 28.5 & - \\
\hline
BEVFormer ~\cite{li2024bevformer} & ResNet-101 & 1600$\times$900  & \ding{55} & 33.7 & - & - & - & 23.7 & 2.4 \\
FB-Occ (16f) ~\cite{li2023fb}  & ResNet-50 & 704$\times$256  & \ding{55} & 35.6 & - & - & - & 27.9 & 9.1 \\
BEVDetOcc (2f) ~\cite{huang2022bevdet4d} & ResNet-50 & 704$\times$256  & \ding{55} & 29.6 & 23.6 & 30.0 & 35.1 & - & 5.5 \\
SparseOcc (8f) ~\cite{liu2024fully} & ResNet-50 & 704$\times$256  & \ding{55} & 34.0 & 28.0 & 34.7 & 39.4 & 30.6 & 22.6 \\
\hline
Panoptic-FlashOcc-tiny (1f) ~\cite{yu2024panoptic} & ResNet-50 & 704$\times$256  & \ding{55} & 34.8 & 29.1 & 35.7 & 39.7 & 29.1 & 45.8 \\
\rowcolor{blue!10}\textbf{HyGE-Occ-tiny (1f) (Ours)} & ResNet-50 & 704$\times$256  & \ding{55} & 36.3 & 30.8 & 37.2 & 41.1 & 29.4 & 37.7 \\
Panoptic-FlashOcc (1f) & ResNet-50 & 704$\times$256  & \ding{55} & 35.2 & 29.4 & 36.0 & 40.1 & 29.4 & 41.9 \\
\rowcolor{blue!10}\textbf{HyGE-Occ (1f) (Ours)} & ResNet-50 & 704$\times$256  & \ding{55} & 36.8 & 31.2 & 37.5 & 41.6 & 29.6 & 33.0 \\
Panoptic-FlashOcc (2f) & ResNet-50 & 704$\times$256  & \ding{55} & 36.8 & 31.2 & 37.6 & 41.5 & 30.3 & \textbf{49.7} \\
\rowcolor{blue!10}\textbf{HyGE-Occ (2f) (Ours)} & ResNet-50 & 704$\times$256  & \ding{55} & 37.8 & 32.2 & 38.6 & 42.6 & 30.4 & 39.2 \\
Panoptic-FlashOcc (8f) & ResNet-50 & 704$\times$256  & \ding{55} & 38.5 & 32.8 & 39.3 & 43.4 & 31.6 & \underline{48.1} \\
\rowcolor{blue!10}\textbf{HyGE-Occ (8f) (Ours)} & ResNet-50 & 704$\times$256  & \ding{55} & 39.9 & 34.5 & 40.7 & 44.5 & 32.0 & 36.8 \\
\hline

ALOcc-2D-mini (16f) ~\cite{chen2025alocc} & ResNet-50 & 704$\times$256 & \ding{55} & 39.3 & 32.9 & 40.1 & 44.8 & 33.4 & 23.6 \\
\rowcolor{blue!10}\quad+\textbf{HyGE-Occ} & ResNet-50 & 704$\times$256 & \ding{55} & 40.2 & 33.9 & 40.9 & 45.6 & 33.9 & 22.8 \\

ALOcc-2D (16f) & ResNet-50 & 704$\times$256 & \ding{55} & \underline{43.0} & \underline{37.1} & \underline{43.8} & \underline{48.2} & \textbf{37.4} & 9.8 \\
\rowcolor{blue!10}\quad+\textbf{HyGE-Occ} & ResNet-50 & 704$\times$256 & \ding{55} & \textbf{43.2} & \textbf{37.3} & \textbf{44.0} & \textbf{48.3} & \underline{37.1} & 9.5 \\
\bottomrule
\end{tabular}
}
\vspace{-.5em}
\caption{Comparison with occupancy and panoptic occupancy prediction models on the Occ3D-nuScenes validation set. HyGE-Occ achieves the best overall performance across RayIoU and mIoU metrics while maintaining competitive efficiency, demonstrating its effectiveness for accurate 3D occupancy prediction. Vis. Mask refers to camera mask}
\vspace{-1em}
\label{tab:main_tab_ray_iou}
\end{table*}

\begin{table}[t]
\centering
\resizebox{\linewidth}{!}{

\begin{tabular}{l|c|c c c}
\toprule
\textbf{Method} & 
\textbf{RayPQ} & \textbf{RayPQ$_{1m}$} & \textbf{RayPQ$_{2m}$} &
\textbf{RayPQ$_{4m}$} \\
\midrule
SparseOcc (8f) ~\cite{liu2024fully}           & 14.1 & 10.2 & 14.5 & 17.6 \\
\hline
Panoptic-FlashOcc-tiny (1f) ~\cite{yu2024panoptic}  & 12.9 &  8.8 & 13.4 & 16.5 \\
\rowcolor{blue!10}\textbf{HyGE-Occ-tiny (1f)} & 14.5 & 10.5 & 14.9 & 18.1 \\
Panoptic-FlashOcc (1f)        & 13.2 &  9.2 & 13.5 & 16.8 \\
\rowcolor{blue!10}\textbf{HyGE-Occ (1f)} & 15.1 & 11.1 & 15.5 & 18.9 \\
Panoptic-FlashOcc (2f)        & 14.5 & 10.6 & 15.0 & 18.0 \\
\rowcolor{blue!10}\textbf{HyGE-Occ (2f)} & 15.8 & 11.7 & 16.1 & 19.5 \\
Panoptic-FlashOcc (8f)        & \underline{16.0} & \underline{11.9} & \underline{16.3} & \underline{19.7} \\
\rowcolor{blue!10}\textbf{HyGE-Occ (8f)} & \textbf{17.5} & \textbf{13.1} & \textbf{17.9} & \textbf{21.4} \\
\bottomrule
\end{tabular}
}
\vspace{-.5em}
\caption{Panoptic quality measured on the Occ3D-nuScenes validation set. HyGE-Occ consistently outperforms prior methods across all RayPQ metrics, demonstrating improved instance coherence and overall panoptic quality in 3D scene understanding.}
\vspace{-1.5em}
\label{tab:main_tab_ray_pq}
\end{table}

The overall training objective jointly optimizes semantic and instance-level predictions:
\begin{equation}\label{occ_loss}
\mathcal{L}_{\text{occ}} =
\lambda_{\text{sem}} \mathcal{L}_{\text{sem}} +
\lambda_{\text{center}} \mathcal{L}_{\text{center}} +
\lambda_{\text{offset}} \mathcal{L}_{\text{offset}}.
\end{equation}

This unified formulation allows HyGE-Occ to achieve both high-fidelity geometric reconstruction and sharp panoptic delineation, while maintaining full compatibility with existing occupancy frameworks.

\subsection{Training Objective}

Our framework is trained end-to-end with four complementary loss components:  
the original discretized LSS loss ($\mathcal{L}_{\text{LSS}}$), the Gaussian loss ($\mathcal{L}_{\text{G}}$), the panoptic occupancy loss ($\mathcal{L}_{\text{occ}}$), and the edge loss ($\mathcal{L}_{\text{edge}}$).  
Each term ensures that the hybrid representation maintains both spatial coverage and boundary precision.
The overall training objective is as:
\begin{equation}\label{total_loss}
    \mathcal{L}_{\text{total}} =
    \lambda_{\text{LSS}} \mathcal{L}_{\text{LSS}} +
    \lambda_{\text{G}} \mathcal{L}_{\text{G}} +
    \lambda_{\text{edge}} \mathcal{L}_{\text{edge}} +
    \mathcal{L}_{\text{occ}},
\end{equation}
where $\lambda_{\text{LSS}}$, $\lambda_{\text{G}}$, and $\lambda_{\text{edge}}$ control the relative contributions of each auxiliary component.  
All modules are trained jointly, enabling the hybrid view-transformation branch to learn complementary geometric cues while the edge supervision refines boundary precision.





\label{sec:Training_obj}
\section{Experiments}
\label{sec:experiments}
\subsection{Experimental Setup}
\textbf{Dataset.} We use the Occ3d-nuScenes~\cite{tian2023occ3d} dataset for all experiments, which provides 700 training, 150 validation and 150 test scenes with multimodal inputs from LiDAR and RGB sensors. The occupancy space spans from -40m to 40m horizontally and from -1 to 5.4m vertically, where each voxel is a cube with 0.4m sides and represents occupancy labels for 17 distinct classes.

\noindent\textbf{Metric.}
Following previous works, we use mIoU, RayIoU, and RayPQ metrics for model evaluation.
The mIoU measures voxel-wise semantic accuracy by averaging the intersection-over-union (IoU) across all semantic categories.
Additionally, RayIoU extends mIoU to a ray-based formulation by casting query rays into the predicted 3D volume and evaluating geometric consistency along the viewing direction within depth thresholds (1m, 2m, 4m).
Furthermore, RayPQ builds upon the Panoptic Quality (PQ) metric, jointly assessing semantic and instance-level consistency along rays, thus providing a comprehensive evaluation for panoptic 3D occupancy prediction. Note that detailed descriptions of implementation details are provided in the supplementary materials.

\subsection{Comparison with SOTA Methods}
\begin{figure}[t]
    \centering
    \includegraphics[width=\linewidth]{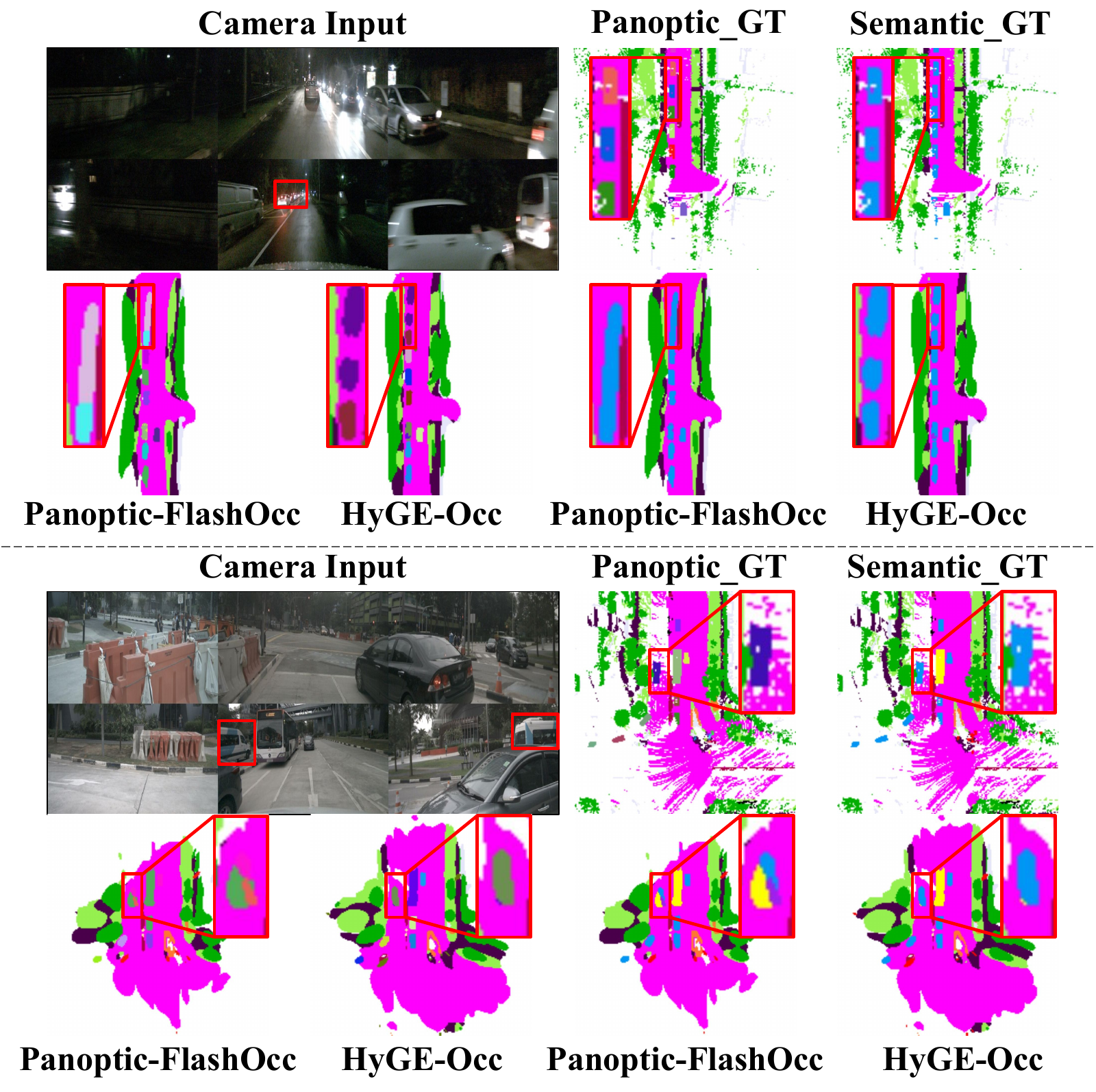}
    \vspace{-2.em}
    \caption{
    Qualitative comparison on the Occ3D-nuScenes validation set. Compared to the baseline Panoptic-FlashOcc, the proposed \textbf{HyGE-Occ} produces more accurate and coherent panoptic occupancy predictions. Our method better delineates semantic and instance boundaries, particularly in regions with dense object interactions and occlusions.
    }
    \label{fig:qual_fig}
    \vspace{-.5em}
\end{figure}

\textbf{Quantitative Results.} We first quantitatively compare our proposed framework with recent state-of-the-art camera-based occupancy prediction methods on the Occ3D-nuScenes validation set. 
While existing approaches have advanced the field through improved geometric reasoning and BEV-based spatial aggregation, most remain limited by discrete depth formulations or insufficient boundary modeling, which can lead to coarse geometry and blurred instance boundaries as seen in Figure~\ref{fig:qual_fig}.
Our hybrid design, combining continuous Gaussian reasoning with discretized LSS features, addresses these issues by enhancing geometric consistency and preserving fine structural details. 
In addition, the edge head explicitly guides the network to recover sharper object contours and more coherent instance separation within the 3D occupancy space.

As shown in Tables~\ref{tab:main_tab_ray_iou} and~\ref{tab:main_tab_ray_pq}, our framework consistently outperforms previous methods in both semantic and panoptic occupancy benchmarks, achieving improvements in geometric fidelity and panoptic quality. Compared to panoptic models such as Panoptic-FlashOcc and SparseOcc, HyGE-Occ demonstrates superior performance across all major metrics. Also, when we applied our method to a semantic model, ALOcc, improvements were shown, validating the effectiveness of the proposed hybrid view-transformation branch and edge head design. These results highlight HyGE-Occ as a new strong baseline for camera-based 3D occupancy prediction, achieving robust and accurate 3D scene understanding in complex environments.

\begin{table}[t]
\centering

\resizebox{\linewidth}{!}{
\begin{tabular}{l|c|c|c c c}

\toprule
\textbf{Method} & \textbf{Hybrid} & \textbf{Edge} & \textbf{RayIoU} & \textbf{mIoU} & \textbf{FPS} \\
\midrule

\multirow{4}{*}{\textbf{HyGE-Occ-tiny (1f)}} 
& \ding{55} & \ding{55} & 34.8 & 29.1 & \textbf{45.8} \\ 
& \ding{51} & \ding{55} & 35.3 & 29.3 & 38.0 \\
& \ding{55} & \ding{51} & 35.4 & 29.3 & 45.7 \\
& \ding{51} & \ding{51} & \textbf{36.3} & \textbf{29.4} & 37.7 \\

\hline
\multirow{4}{*}{\makecell[l]{ALOcc-2D-mini (16f) ~\cite{chen2025alocc}\\+ \textbf{HyGE-Occ}}}
& \ding{55} & \ding{55} & 39.3 & 33.4 & \textbf{23.6} \\ 
& \ding{51} & \ding{55} & 39.7 & 33.6 & 22.9 \\
& \ding{55} & \ding{51} & 39.7 & 33.5 & 23.5 \\
& \ding{51} & \ding{51} & \textbf{40.2} & \textbf{33.9} & 22.8 \\

\bottomrule

\end{tabular}
}
\vspace{-.5em}
\caption{Ablation study to investigate the effect of our two modules. The effectiveness of each module in HyGE-Occ is evaluated by selectively enabling the hybrid view-transformation branch and the edge head. Both modules contribute to higher RayIoU and mIoU, with the best performance achieved when combined.}
\vspace{-1.5em}
\label{tab:ablation_tab}
\end{table}


\noindent\textbf{Qualitative Results.} We provide qualitative comparisons in Figure.~\ref{fig:qual_fig} to further demonstrate the effectiveness of our proposed framework. 
Compared to the Panoptic-FlashOcc model, HyGE-Occ produces more precise object boundaries and more coherent semantic occupancy maps. Also our model shows better performance in distinguishing between different instances. 
The edge head sharpens inter-instance borders, reducing label bleeding and improving panoptic separation. 
The hybrid view-transformation branch enhances geometric continuity and spatial coherence by alleviating discretization artifacts commonly observed in depth-based unprojection. 
This continuous and uncertainty-aware reasoning results in smoother scene geometry, fewer fragmented regions, and more stable semantic predictions across complex environments. Together, these visual improvements confirm that our framework effectively enhances both geometric fidelity and instance-level delineation within the 3D occupancy space.

\begin{figure}[t]
    \centering
    \includegraphics[width=\linewidth]{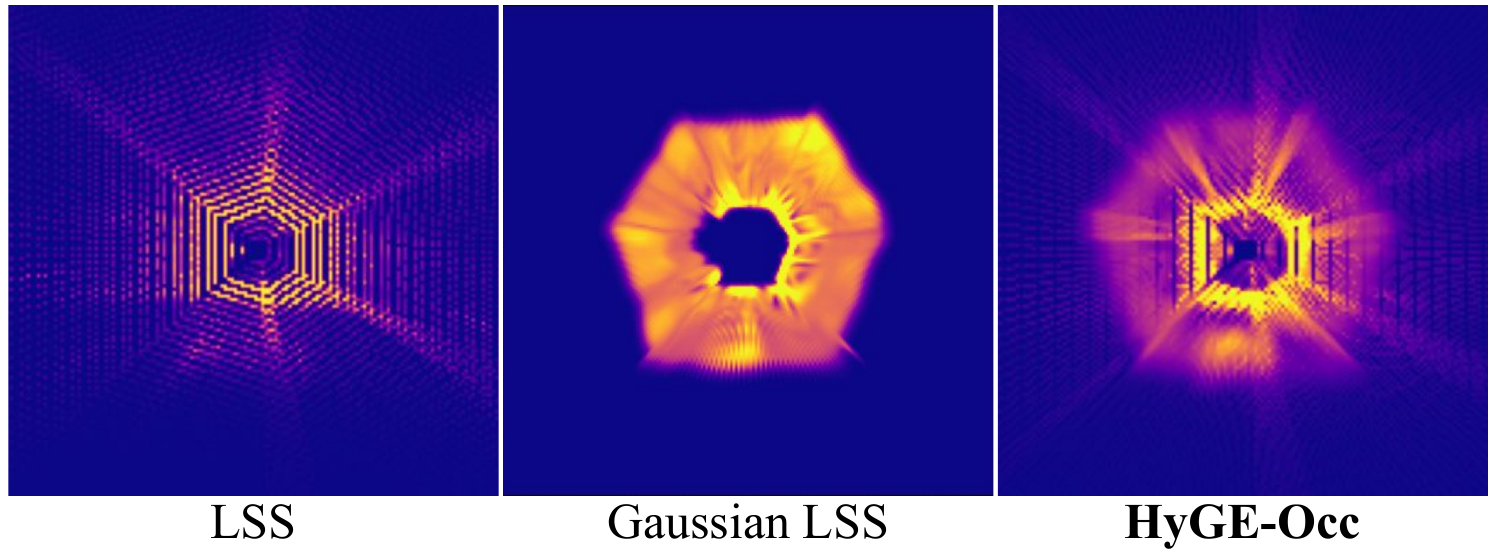}
    \vspace{-1.5em}
    \caption{
    Visualization of BEV features. Comparison between BEV features generated by the discretized LSS, continuous GaussianLSS, and our proposed hybrid representation. 
    }
    \label{fig:bev_fig}
    \vspace{-.5em}
\end{figure}



\subsection{Ablation Studies}
\begin{table}[t]
\centering
\footnotesize
\begin{tabular}{l|c|c}

\toprule

\textbf{Method} & \textbf{Gating} & \textbf{Alpha-Blending} \\
\midrule
\textbf{RayIoU} / \textbf{mIoU} & 36.3 / 29.1 & \textbf{36.3} / \textbf{29.4} \\
\bottomrule

\end{tabular}
\vspace{-.5em}
\caption{Ablation study on feature fusion. Comparison between gating and $\alpha$-blending strategies for merging discrete and continuous BEV features. Both achieve comparable results, indicating that simple $\alpha$-blending provides sufficient fusion effectiveness.}
\vspace{-.5em}
\label{tab:feat_fusion_tab}
\end{table}
\begin{table}[t]
\centering
\footnotesize
\resizebox{\linewidth}{!}{
\begin{tabular}{l|ccccc}

\toprule
\textbf{Method} & \textbf{0.2} & \textbf{0.4} & \textbf{0.6} & \textbf{0.8} & \textbf{1.0} \\
\midrule
\textbf{HyGE-Occ-tiny (1f)} & 36.2 & 36.2 & \textbf{36.3} & 36.2 & 36.1\\ 
ALOcc-2D-mini (16f) + \textbf{HyGE-Occ} & 39.5 & 40.0 & \textbf{40.2} & 39.7 & 39.6 \\ 
\bottomrule

\end{tabular}
}
\vspace{-.5em}
\caption{Ablation study on the $\alpha$-blending coefficient for hybrid feature fusion. The RayIoU remains stable across different $\alpha$ values, indicating that HyGE-Occ is robust to the fusion weight.}
\vspace{-.5em}
\label{tab:alpha_tab}
\end{table}

\textbf{Effect of Individual Modules.}~
Table~\ref{tab:ablation_tab} presents quantitative ablation results on the Occ3D-nuScenes validation set using Panoptic-FlashOcc (1f) and ALOcc-2D-mini as baselines. The hybrid branch consistently improves geometric prediction quality by integrating continuous and discrete depth reasoning, while the edge prior further enhances local boundary precision and instance separation.  
When the two modules are combined, the model achieves the best overall performance, demonstrating their complementary nature—where the hybrid branch strengthens global structural consistency and the edge prior refines fine-grained contours and object boundaries across varied scene conditions.

\begin{figure}[t]
    \centering
    \includegraphics[width=\linewidth]{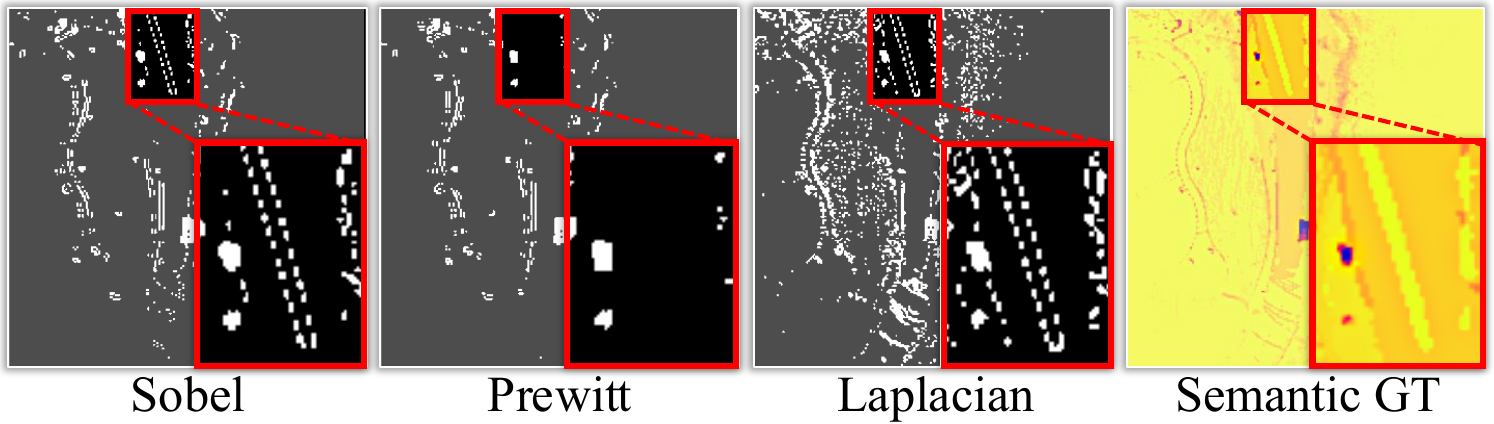}
    \vspace{-1.5em}
    \caption{
    \textbf{Comparison of edge kernels.} Edge maps are generated using Sobel, Prewitt, and Laplacian operators. Each are compared with the semantic ground truth. The Sobel operator produces sharper and more consistent boundaries with less noise, making it the most effective choice for our edge supervision.
    }
    \label{fig:edge_fig}
    \vspace{-1.5em}
\end{figure}

\noindent\textbf{Effect of the Hybrid View-Transformation Branch.}~
To evaluate the impact of the proposed hybrid view-transformation design, we compare BEV representations obtained from LSS, GaussianLSS, and our hybrid branch.  
As shown in Figure~\ref{fig:bev_fig}, the difference between LSS and GaussianLSS is clearly visible:  
LSS produces discretized, discontinuous BEV features due to its bin-based depth reasoning, leading to aliasing artifacts around object boundaries.  
In contrast, GaussianLSS models depth as a continuous distribution, producing smoother and geometrically coherent spatial features.  
By blending the two through our hybrid view-transformation branch, HyGE-Occ achieves both geometric stability and fine-grained spatial fidelity, resulting in a more consistent BEV representation.

We further analyze different feature fusion strategies in Table~\ref{tab:feat_fusion_tab}. In addition to simple alpha-blending, we evaluate a gating function that predicts a spatially varying fusion weight map, allowing the network to adaptively emphasize either the continuous or discretized features at each BEV location. Although this gating mechanism provides greater flexibility, it introduces additional parameters and increases training sensitivity. In contrast, alpha-blending yields slightly higher overall accuracy while maintaining a simpler and more stable formulation. This indicates that a simple linear fusion is sufficient to integrate the complementary cues from continuous and discretized depth reasoning without requiring complex adaptive weighting schemes.


We also examine the effect of the blending coefficient $\alpha$ in Table~\ref{tab:alpha_tab}. Across both models, performance remains stable for a wide range of $\alpha$ values, demonstrating that the hybrid representation is robust to the choice of blending ratio. The performance gradually improves as $\alpha$ grows from 0.2 to 0.6, indicating that incorporating stronger Gaussian cues enhances geometric consistency and feature smoothness.
Beyond $\alpha=0.6$, the gain saturates or slightly declines, suggesting that excessive reliance on Gaussian features may suppress the fine-grained local details captured by the discretized LSS branch.
Overall, $\alpha=0.6$ provides the best trade-off between geometric fidelity and spatial precision across different frameworks, offering a stable default choice for integrating the two complementary representations.
This suggests that the hybrid formulation effectively leverages complementary geometric cues from both representations without requiring fine-grained tuning.

\noindent\textbf{Effect of the Edge Prior Module.}~
To examine the impact of the proposed edge prior, we conduct ablative experiments on the choice of edge extraction kernels and their corresponding hyperparameters.  
The pseudo edge labels are generated by applying standard gradient-based operators to ground-truth semantic maps, as visualized in Figure~\ref{fig:edge_fig}.  
Among the tested filters, the Sobel operator provides the most stable and well-localized boundaries, which translate into high-quality pseudo edge supervision.
Table~\ref{tab:kernel_tab} quantitatively confirms this observation.  
Models trained with Sobel-based pseudo edge labels achieve the best performance (\textbf{36.3} / \textbf{29.4}) compared to Prewitt~\cite{prewitt1970object} (36.3 / 29.1) and Laplacian~\cite{paris2011local} (36.0 / 29.2),  
demonstrating that accurate edge localization is critical for effective boundary guidance.  
We further analyze the effect of kernel size $S_{x,y}$ and weighting factor $\lambda_{\text{edge}}$ in Table~\ref{tab:edge_ablation}.  
The results indicate that a $3 \times 3$ kernel with $\lambda_{\text{edge}}=4.0$ yields the best trade-off between boundary sharpness and global structural consistency.  
Larger kernels ($5 \times 5$ or $7 \times 7$) introduce redundant edge noise and slightly degrade performance, while smaller or overly strong weights reduce overall stability.  
Consequently, we adopt the Sobel operator with a $3 \times 3$ kernel and $\lambda_{\text{edge}}=4.0$ as the default experiment configuration.

\begin{table}[t]
\centering
\footnotesize
\begin{tabular}{l|c|c|c}
\toprule
\textbf{Method} & \textbf{Sobel} & \textbf{Prewitt} & \textbf{Laplacian} \\ 
\midrule
\textbf{RayIoU} / \textbf{mIoU} & \textbf{36.3} / \textbf{29.4} & 36.3 / 29.1 & 36.0 / 29.2 \\
\bottomrule

\end{tabular}
\vspace{-.5em}
\caption{Ablation study on edge detection kernels. Different edge detection kernels are compared for generating pseudo edge labels. The Sobel operator yields the best performance, indicating its effectiveness in capturing clear and consistent structural boundaries.}
\vspace{-.5em}
\label{tab:kernel_tab}
\end{table}

\begin{table}[t]
\centering
\footnotesize
\begin{tabular}{c|c|c|c}

\toprule
\diagbox{$S_{x,y}$}{$\lambda_\text{edge}$} & 2.0 & 4.0 & 8.0 \\
\midrule
3 $\times$ 3 & 36.3 / 29.3 & \textbf{36.3} / \textbf{29.4} & 36.1  / 29.2 \\ 
5 $\times$ 5 & 36.3 / 29.3 & 36.3 / 29.1 & 36.2 / 29.2 \\
7 $\times$ 7 & 36.3 / 29.1 & 36.1 / 29.1 & 36.2 / 29.2 \\
\bottomrule

\end{tabular}
\vspace{-.5em}
\caption{Ablation study on edge loss weight ($\lambda_{\text{edge}}$) and Sobel kernel size ($S_{x,y}$). The results show that moderate edge supervision ($\lambda_{\text{edge}}=4.0$) with a $3\times3$ kernel achieves the best balance between geometric accuracy and boundary precision.}
\vspace{-1.5em}
\label{tab:edge_ablation}
\end{table}

\section{Conclusion}
\label{sec:conclusion}


We presented HyGE-Occ, a hybrid view-transformation framework with 3D Gaussian and edge priors that enhances geometric consistency and boundary precision in 3D panoptic occupancy prediction. By combining continuous 3D Gaussian based reasoning with discretized depth features and introducing an edge-guided BEV refinement, HyGE-Occ produces sharper instance delineation and more coherent 3D geometry. Evaluated on the Occ3D-nuScenes benchmark, it achieves state-of-the-art performance, surpassing strong baselines such as Panoptic-FlashOcc and ALOcc.

\noindent \textbf{Limitations.}~HyGE-Occ is limited to BEV-based models, and its effectiveness may not directly generalize to other architectures. In addition, the hybrid view-transformation branch introduces modest computational overhead compared to discretized LSS-based methods. Exploring more efficient formulations, broader architectural compatibility, and extending to flow prediction remains as future work.

\clearpage


{
    \small
    \bibliographystyle{ieeenat_fullname}
    \bibliography{main}
}


\end{document}


\maketitlesupplementary

\setcounter{page}{1}
\setcounter{section}{0}
\renewcommand{\thesection}{\Alph{section}}

\newcommand{\myparagraph}[1]{\vspace{2pt}\noindent{\bf #1}}
\newcommand\framework{{HyGE-Occ}}

\paragraph{Overview.}

In this supplementary material, we first provide the implementation details of our Hybrid View-Transformation Branch and Edge Prediction Module. Next, we provide experiment settings for training HyGE-Occ and implementing our method to ALOcc model. Then, we present additional qualitative results that further demonstrate the superiority of our framework and the effectiveness of each module. Finally, we present each per-class score for each metric.

\section{Implementation Details}
Our framework is built upon MMCV and PyTorch. We provide the experimental setup for HyGE-Occ and ALOcc and 4 H100 Nvidia GPUs are used for all experiments.

\noindent\textbf{Occ3D-nuScenes Dataset.}
The Occ3D-nuScenes benchmark~\cite{tian2023occ3d} extends the original nuScenes dataset by providing dense 3D voxelized occupancy annotations tailored for camera-based 3D scene understanding. It follows the standard nuScenes split of 700 training, 150 validation, and 150 test scenes, and includes synchronized six-camera RGB inputs with full intrinsic and extrinsic calibration.

\noindent\textbf{Voxel Grid.}
The 3D space around the ego vehicle is discretized into a regular voxel grid used for occupancy prediction. By default, the official Occ3D-nuScenes protocol defines a horizontal coverage of $[-40,40]$\,m and a vertical range of $[-1,5.4]$\,m, with a voxel resolution of $0.4$\,m in $x$ and $y$ and $0.4$\,m in $z$. Each voxel is assigned one of 17 semantic classes, along with instance IDs for foreground objects, enabling panoptic 3D occupancy evaluation.  

\noindent\textbf{Ray-based Evaluation.}
Occ3D-nuScenes evaluates camera-only models using ray-level metrics. For each camera pixel, a ray is cast into the 3D scene. A prediction is considered correct if the semantic label matches the ground truth and the predicted depth is within a tolerance of 1\,m, 2\,m, or 4\,m. These ray-wise true positives, false positives, and false negatives form the basis of the RayIoU and RayPQ metrics used throughout our experiments.

\noindent\textbf{Hyperparameter values.}
The continuous Gaussian-based unprojection branch is optimized with 
\begin{equation}
\mathcal{L}_{\text{G}} = \lambda_1 \mathcal{L}_{\text{seg}} + \lambda_2 \mathcal{L}_{\text{center}\_g} + \lambda_3 \mathcal{L}_{\text{offset}\_g},
\end{equation} and the loss weights are set as follows. The segmentation loss is weighted by $\lambda_{1}=5.0$, the centerness loss by $\lambda_{2}=10.0$, and the offset loss by $\lambda_{3}=0.5$, reflecting the relative importance of each component in the continuous Gaussian branch.

The panoptic occupancy head is optimized with
\begin{equation}\label{occ_loss}
\mathcal{L}_{\text{occ}} =
\lambda_{\text{sem}} \mathcal{L}_{\text{sem}} +
\lambda_{\text{center}} \mathcal{L}_{\text{center}} +
\lambda_{\text{offset}} \mathcal{L}_{\text{offset}},
\end{equation}
and the loss weights are set uniformly to $\lambda_{\text{sem}}=\lambda_{\text{center}}=\lambda_{\text{offset}}=100$.

The overall training objective combines the losses from the discretized LSS branch, the continuous Gaussian branch, the edge prediction module, and the panoptic occupancy head.  
Formally, the total loss is calculated with \begin{equation}\label{total_loss}
    \mathcal{L}_{\text{total}} =
    \lambda_{\text{LSS}} \mathcal{L}_{\text{LSS}} +
    \lambda_{\text{G}} \mathcal{L}_{\text{G}} +
    \lambda_{\text{edge}} \mathcal{L}_{\text{edge}} +
    \mathcal{L}_{\text{occ}},
\end{equation} where $\lambda_{\text{LSS}}$, $\lambda_{\text{G}}$, and $\lambda_{\text{edge}}$ control the relative contributions of the LSS loss, Gaussian branch loss, and edge loss, respectively.  
Unless otherwise specified, all loss weights are set to $1$ for stable and balanced training. 

We train HyGE-Occ for a total of 48 epochs using the AdamW optimizer~\cite{loshchilov2017decoupled} with an initial learning rate of $10^{-4}$ and a weight decay of $10^{-2}$. The learning rate follows a step schedule with a linear warm-up of 200 iterations and a warm-up ratio of $0.001$, matching the EpochBasedRunner configuration in MMDetection3D. We also adopt an EMA update scheme (MEGVIIEMAHook) with $21{,}120$ initial updates, and evaluate the model every epoch after the 20th epoch using the official validation protocol.

For experiments where HyGE-Occ is applied to the ALOcc backbone, we train with 1 sample per GPU for 24 epochs on the Occ3D-nuScenes training split. We use the AdamW optimizer with an initial learning rate of $2\times10^{-4}$ and weight decay of $10^{-2}$. The learning rate follows a step schedule with a linear warm-up for the first 200 iterations, a warm-up ratio $0.001$, and a single decay step at the end of training. It is implemented with the \texttt{IterBasedRunner} whose maximum iteration count is set to \texttt{num\_epochs}~$\times$~\texttt{num\_iters\_per\_epoch}. We keep the original EMA hook configuration and evaluate using the official validation pipeline.

\section{Metric Details}
We adopt three standard metrics for 3D occupancy and panoptic occupancy evaluation, mIoU, RayIoU, RayPQ. While mIoU measures voxel-level semantic occupancy, RayIoU and RayPQ evaluate predictions along camera rays, which better capture long-range consistency and instance structure than voxel-only metrics. We summarize the definitions below.

\noindent \textbf{Mean Intersection-over-Union (mIoU)}
For semantic occupancy prediction, the IoU of class $c$ is defined as:
\begin{equation}
\mathrm{IoU}_c = 
\frac{|\hat{O}_c \cap O_c|}
     {|\hat{O}_c \cup O_c|},
\end{equation}
The final mIoU is computed by averaging IoU across all semantic classes:
\begin{equation}
\mathrm{mIoU} = 
\frac{1}{C} \sum_{c=1}^{C} \mathrm{IoU}_c,
\end{equation}
where $C$ is the number of classes.

\noindent \textbf{Ray Intersection over Union (RayIoU).}
We follow SparseOcc~\cite{liu2024fully}, a predicted ray is considered a true positive (TP) if its semantic class matches the ground-truth class and the L1 depth error between the prediction and the ground truth falls within a distance threshold $d \in \{1\text{m},2\text{m},4\text{m}\}$. Let $C$ be the number of semantic classes. For each class $c$, let $\mathrm{TP}_c$, $\mathrm{FP}_c$, and $\mathrm{FN}_c$ denote the number of true positives, false positives, and false negatives, respectively. The RayIoU at threshold $d$ is defined as:
\begin{equation}
\mathrm{RayIoU}_d = \frac{1}{C} \sum_{c=1}^{C} \frac{\mathrm{TP}_c}{ \mathrm{TP}_c + \mathrm{FP}_c + \mathrm{FN}_c}.
\end{equation}
The final RayIoU score averages over all thresholds as:
\begin{equation}
\mathrm{RayIoU} = \frac{1}{3} \sum_{d \in \{1,2,4\}} \mathrm{RayIoU}_d.
\end{equation}

\noindent\textbf{Ray Panoptic Quality (RayPQ).}
We follow the setting of SparseOcc~\cite{liu2024fully}, where RayPQ is defined analogously to the standard panoptic quality (PQ), as the product of \emph{segmentation quality} (SQ) and \emph{recognition quality} (RQ)
along rays.

For a given distance threshold $d \in \{1\text{m},2\text{m},4\text{m}\}$, $TP_d$, $FP_d$, and $FN_d$ denote the sets of true-positive, false-positive, and false-negative ray-level matches,  respectively. Each element in $TP_d$ is a matched pair $(p,g)$ of predicted and ground-truth ray instances.

\noindent The segmentation quality at threshold $d$ is measured as:
\begin{equation}
\mathrm{SQ}_d = \frac{1}{|TP_d|} \sum_{(p,g)\in TP_d} \mathrm{IoU}(p,g),
\end{equation}
and the recognition quality is measured as:
\begin{equation}
\mathrm{RQ}_d = \frac{|TP_d|} {|TP_d| + \tfrac{1}{2}|FP_d| + \tfrac{1}{2}|FN_d|}.
\end{equation}

\noindent~Ray panoptic quality at threshold $d$ is then computed by:
\begin{equation}
\mathrm{RayPQ}_d = \mathrm{SQ}_d \times \mathrm{RQ}_d.
\end{equation}

\noindent~Finally, we report the averaged RayPQ over all thresholds:
\begin{equation}
\mathrm{RayPQ} = \frac{1}{3} \sum_{d \in \{1,2,4\}} \mathrm{RayPQ}_d.
\end{equation}

\section{More Qualitative Results.}

\begin{figure*}[h]
    \centering
    \includegraphics[width=\linewidth]{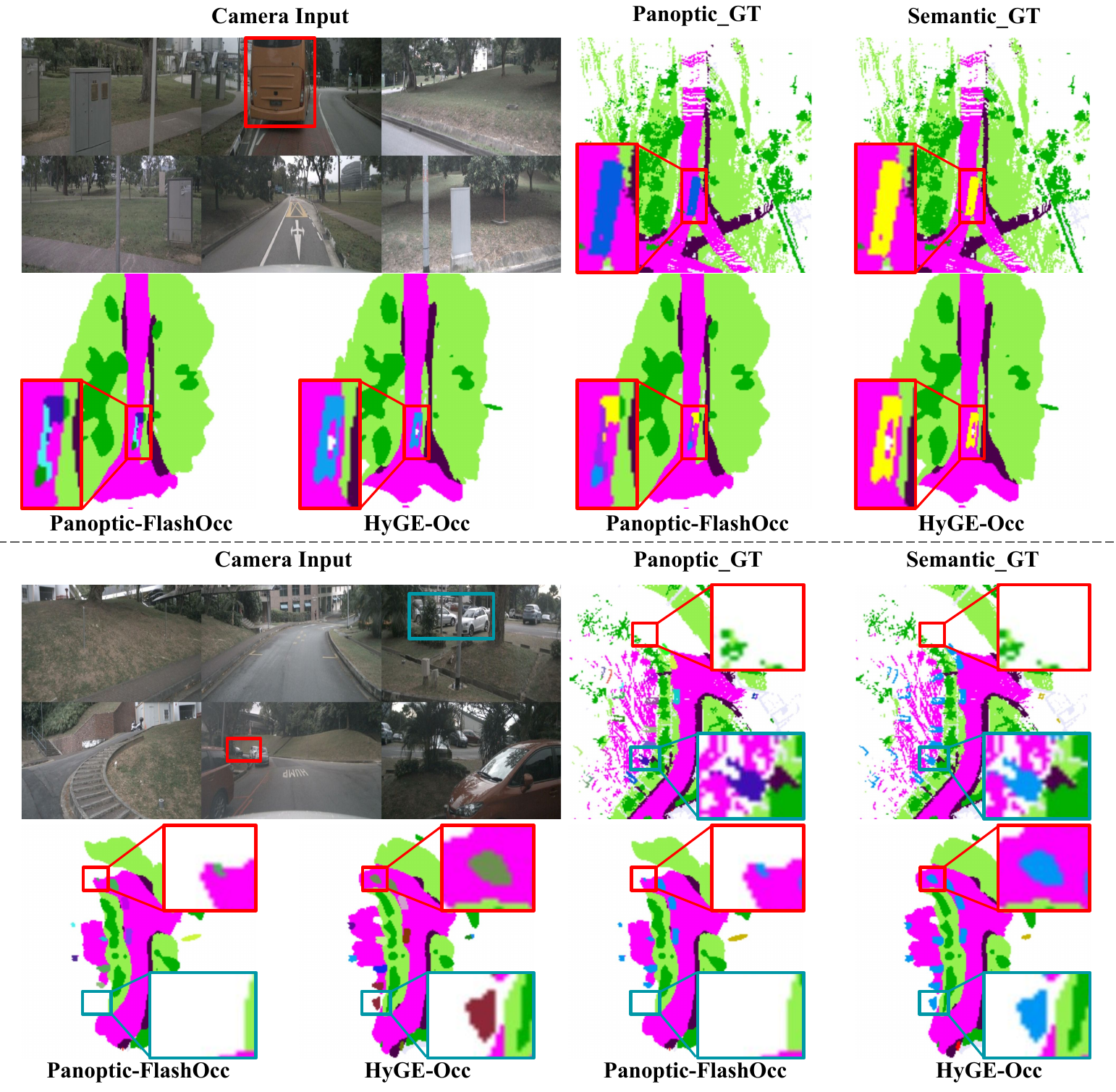}
    \vspace{-2.em}
    \caption{
    Additional qualitative results comparing HyGE-Occ and Panoptic-FlashOcc.}
    \label{fig:supp_qual_HyGE}
    \vspace{-.5em}
\end{figure*}

\begin{figure*}[h]
    \centering
    \includegraphics[width=\linewidth]{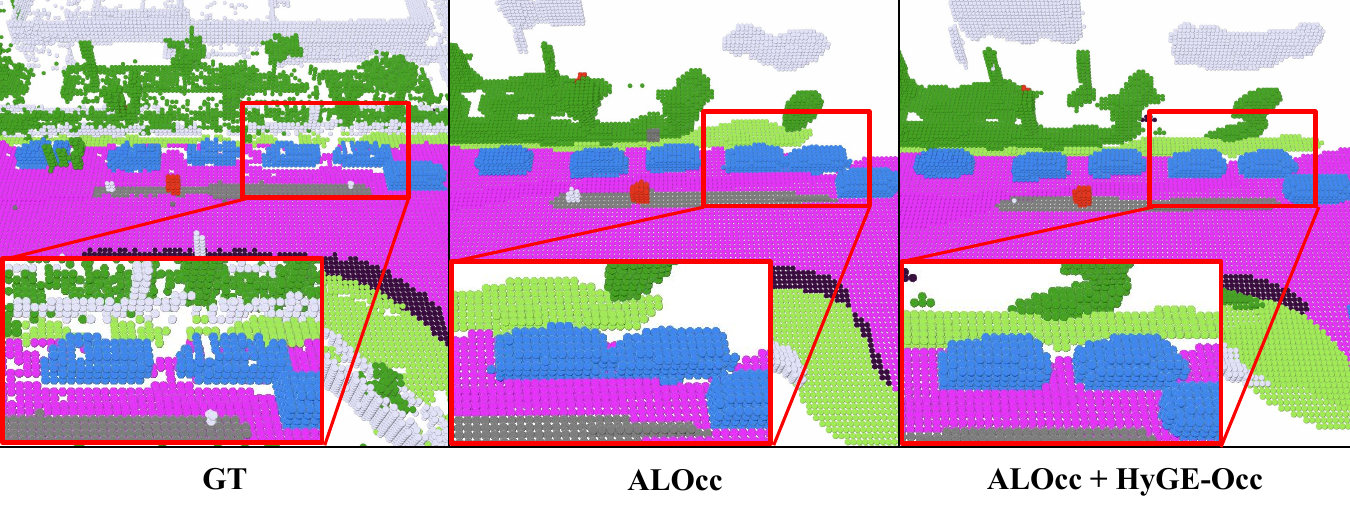}
    \vspace{-2.5em}
    \caption{
    Additional qualitative results comparing ALOcc with HyGE-OCC implemented ALOcc.
    }
    \label{fig:supp_qual_alocc}
    \vspace{-.5em}
\end{figure*}

\noindent\textbf{Qualitative Results.}
We provide additional qualitative comparisons to demonstrate the effectiveness of our proposed framework. As shown in Figure \ref{fig:supp_qual_HyGE}. HyGE-Occ produces more precise object boundaries and more coherent semantic occupancy maps. Also our model shows better performance in distinguishing between different instances. Figure~\ref{fig:supp_qual_alocc} presents comparisons among the semantic ground-truth labels, the original ALOcc model, and ALOcc augmented with HyGE-Occ. When our modules are applied, the model more clearly differentiates object extents and delivers noticeably improved semantic and instance-level delineation.

\begin{figure*}[h]
    \centering
    \includegraphics[width=\linewidth]{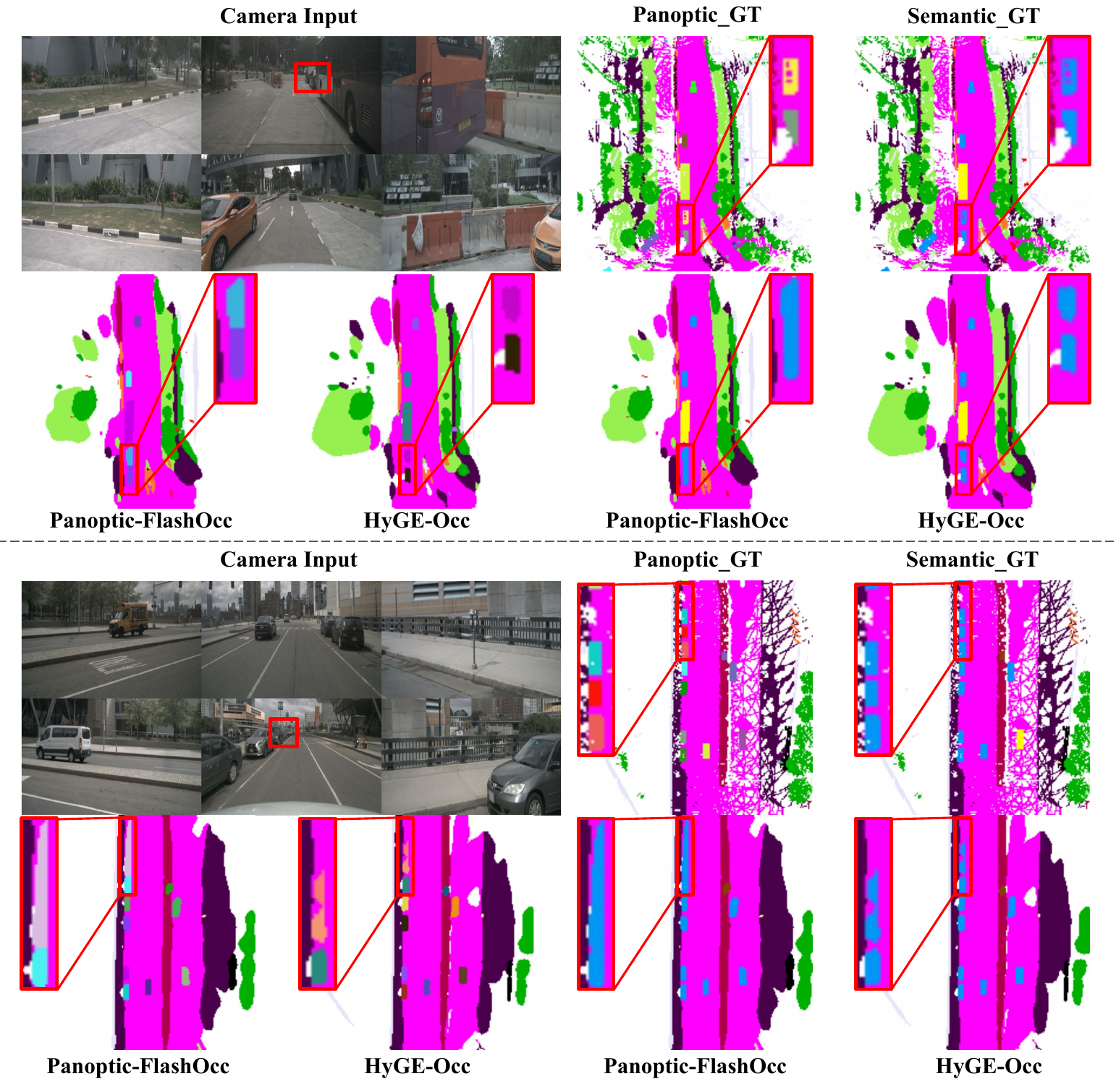}
    \vspace{-2.em}
    \caption{
    Qualitative results from the ablation study on the edge module.
    }
    \label{fig:supp_qual_edge}
    \vspace{-.5em}
\end{figure*}

\begin{figure*}[h]
    \centering
    \includegraphics[width=\linewidth]{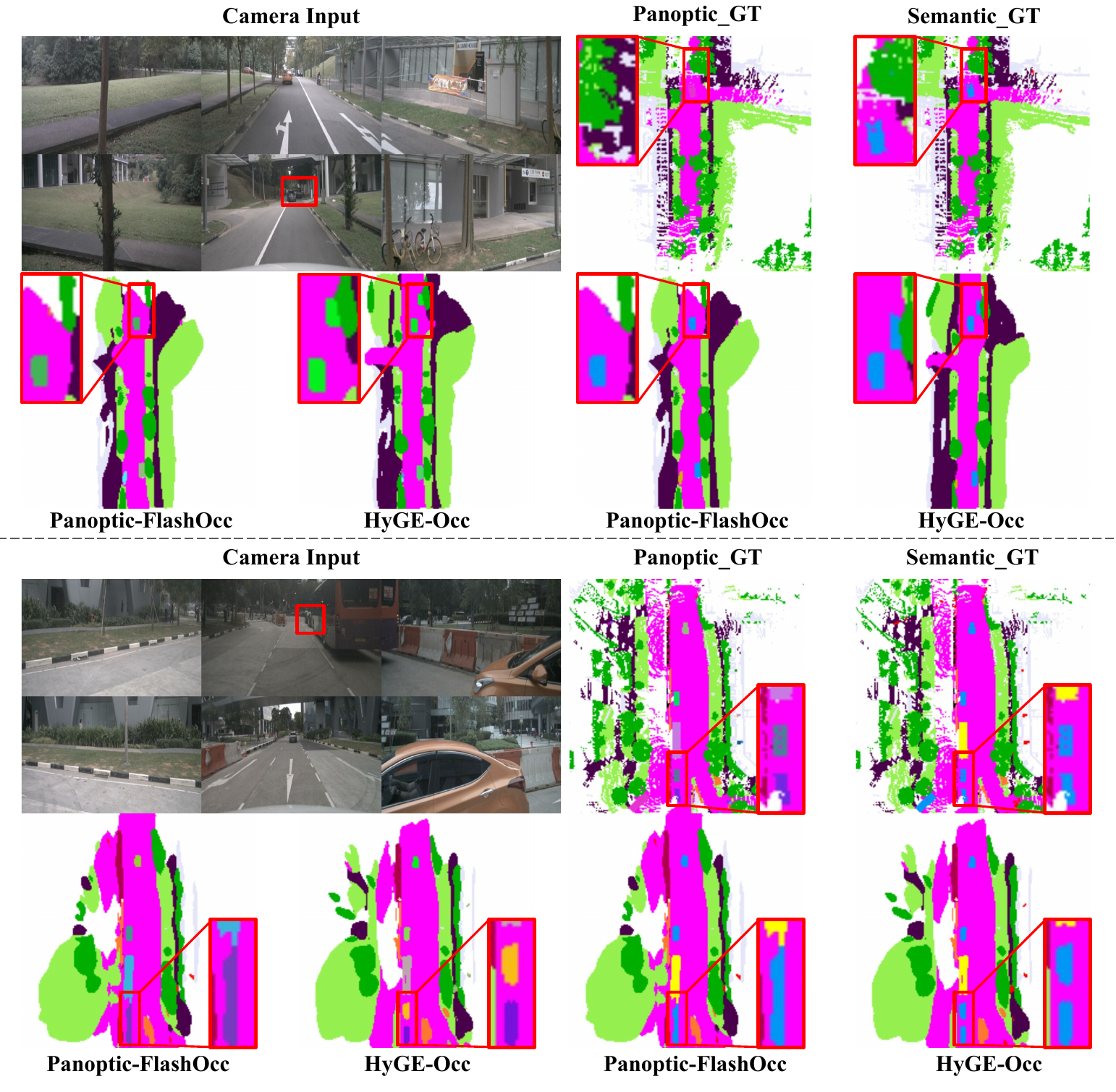}
    \vspace{-2.em}
    \caption{
    Qualitative results from the ablation study on the hybrid view-transformation module.
    }
    \label{fig:supp_qual_gs}
    \vspace{-.5em}
\end{figure*}

\noindent\textbf{Qualitative Results of Ablation Studies.}
We provide additional qualitative comparisons to show the effect of each module when applied independently. As shown in Figure~\ref{fig:supp_qual_edge}, when only the Edge Prediction Module is added, the model produces noticeably sharper object contours and clearer separation between adjacent instances, reducing boundary bleeding and improving instance delineation. In contrast, as shown in Figure \ref{fig:supp_qual_gs}, when only the Hybrid View-Transformation Branch is enabled, the model demonstrates improved geometric stability in depth-ambiguous regions, particularly for distant or partially occluded objects. This is because continuous Gaussian cues help smooth depth transitions and resolve uncertainty in the lifted 3D representation. These visual results confirm that the edge module mainly strengthens boundary precision, while the hybrid branch primarily enhances geometric consistency in challenging depth regions.

\section{Per Class IoU.}
In this section, we present per-class performance for each distance threshold @$d$ and for each evaluation metric. Table~\ref{tab:per_miou} reports the per-class IoU of all methods. HyGE-Occ achieves notably higher IoU on several key categories, including cars, buses, and manmade objects, demonstrating its improved ability to capture object geometry in dense urban scenes. Table~\ref{tab:per_rayiou} shows the per-class RayIoU results. Across all categories, HyGE-Occ consistently outperforms Panoptic-FlashOcc, indicating its stronger geometric consistency along camera rays. When applied to ALOcc, our method also yields clear improvements on bicycle and traffic-cone, highlighting its effectiveness in predicting occupancy for small and thin structures. Finally, Table~\ref{tab:per_raypq} presents the per-class RayPQ scores. HyGE-Occ achieves the highest RayPQ across all categories, confirming that our approach enhances both semantic accuracy and instance consistency in panoptic occupancy prediction.

\begin{table*}[t]
\centering
\resizebox{\textwidth}{!}{
\begin{tabular}{l|c|c|c|c|ccc|c}
\toprule
\textbf{Method} & \textbf{Backbone} & \textbf{Input Size} & \textbf{Vis. Mask} & 
\textbf{RayIoU} & \textbf{RayIoU$_{1m}$} & \textbf{RayIoU$_{2m}$} & \textbf{RayIoU$_{4m}$} & 
\textbf{mIoU} \\
\midrule
MonoScene ~\cite{cao2022monoscene}  & ResNet-101 & 928$\times$1600  & \ding{51} & - & - & - & - & 6.0 \\
TPVFormer ~\cite{huang2023tri}  & ResNet-101 & 928$\times$1600  & \ding{51} & - & - & - & - & 27.8 \\
RenderOcc ~\cite{pan2024renderocc}   & Swin-Base & 512$\times$1408  & \ding{51} & - & - & - & - & 26.1 \\
BEVFormer ~\cite{li2024bevformer} & ResNet-101 & 1600$\times$900  & \ding{51} & 32.4 & 26.1 & 32.9 & 38.0 & 39.3 \\
FB-Occ (16f) ~\cite{li2023fb} & ResNet-50 & 704$\times$256  & \ding{51} & 33.5 & 26.7 & 34.1 & 39.7 & 39.1 \\
FlashOcc ~\cite{yu2023flashocc} & ResNet-50 & 704$\times$256  & \ding{51} & - & - & - & - & 32.0 \\
\hline
RenderOcc ~\cite{pan2024renderocc} & Swin-Base & 1408$\times$512  & \ding{55} & 19.5 & 13.4 & 19.6 & 25.5 & - \\
BEVDetOcc (8f) & ResNet-50 & 704$\times$384  & \ding{55} & 32.6 & 26.6 & 33.1 & 38.2 & 39.3 \\
SparseOcc (16f) & ResNet-50 & 704$\times$256  & \ding{55} & 35.1 & 29.1 & 35.8 & 40.3 & 30.9 \\
\hline
Panoptic-FlashOcc-tiny (1f) ~\cite{yu2024panoptic} & ResNet-50 & 704$\times$256  & \ding{55} & 34.8 & 29.1 & 35.7 & 39.7 & 29.1 \\
\rowcolor{blue!10}\textbf{HyGE-Occ-tiny (1f) (Ours)} & ResNet-50 & 704$\times$256  & \ding{55} & 36.3 & 30.8 & 37.2 & 41.1 & 29.4 \\
Panoptic-FlashOcc (1f) & ResNet-50 & 704$\times$256  & \ding{55} & 35.2 & 29.4 & 36.0 & 40.1 & 29.4 \\
\rowcolor{blue!10}\textbf{HyGE-Occ (1f) (Ours)} & ResNet-50 & 704$\times$256  & \ding{55} & 36.8 & 31.2 & 37.5 & 41.6 & 29.6 \\
Panoptic-FlashOcc (2f) & ResNet-50 & 704$\times$256  & \ding{55} & 36.8 & 31.2 & 37.6 & 41.5 & 30.3 \\
\rowcolor{blue!10}\textbf{HyGE-Occ (2f) (Ours)} & ResNet-50 & 704$\times$256  & \ding{55} & 37.8 & 32.2 & 38.6 & 42.6 & 30.4 \\
Panoptic-FlashOcc (8f) & ResNet-50 & 704$\times$256  & \ding{55} & 38.5 & 32.8 & 39.3 & 43.4 & 31.6 \\
\rowcolor{blue!10}\textbf{HyGE-Occ (8f) (Ours)} & ResNet-50 & 704$\times$256  & \ding{55} & 39.9 & 34.5 & 40.7 & 44.5 & 32.0 \\

\hline
ALOcc-2D-mini (16f) ~\cite{chen2025alocc} & ResNet-50 & 704$\times$256 & \ding{55} & 39.3 & 32.9 & 40.1 & 44.8 & 33.4 \\
\rowcolor{blue!10}\quad+\textbf{HyGE-Occ} & ResNet-50 & 704$\times$256 & \ding{55} & 40.2 & 33.9 & 40.9 & 45.6 & 33.9 \\

ALOcc-2D (16f) & ResNet-50 & 704$\times$256 & \ding{55} & \underline{43.0} & \underline{37.1} & \underline{43.8} & \underline{48.2} & \textbf{37.4} \\
\rowcolor{blue!10}\quad+\textbf{HyGE-Occ} & ResNet-50 & 704$\times$256 & \ding{55} & \textbf{43.2} & \textbf{37.3} & \textbf{44.0} & \textbf{48.3} & \underline{37.1} \\
\bottomrule
\end{tabular}
}
\vspace{-.5em}
\caption{Additional quantitative results comparing with other occupancy and panoptic occupancy prediction models.}
\vspace{-1em}
\label{tab:supp_main}
\end{table*}


\begin{table*}[t]
\centering
\scriptsize
\setlength{\textwidth}{2.8pt}

\newcommand{\clsbox}[1]{\colorbox{#1}{\rule{0pt}{1pt}\rule{1pt}{0pt}}}

\definecolor{cbarrier}{HTML}{FF7832}
\definecolor{cbicycle}{HTML}{FFC0CB}
\definecolor{cbus}{HTML}{FFFF00}
\definecolor{ccar}{HTML}{0096F5}
\definecolor{cconsveh}{HTML}{00FFFF}
\definecolor{cmotor}{HTML}{C8B400}
\definecolor{cped}{HTML}{FF0000}
\definecolor{ctfcone}{HTML}{FFF096}
\definecolor{ctrailer}{HTML}{873C00}
\definecolor{ctruck}{HTML}{A020F0}
\definecolor{cdrvsurf}{HTML}{FF00FF}
\definecolor{cotherflat}{HTML}{AF004B}
\definecolor{csidewalk}{HTML}{4B004B}
\definecolor{cterrain}{HTML}{96F050}
\definecolor{cmannmade}{HTML}{E6E6FA}
\definecolor{cveg}{HTML}{00AF00}

\resizebox{\linewidth}{!}{
\begin{tabular}{l|c|c|c|
                *{16}{>{\centering\arraybackslash}p{1.1em}}}
\toprule
\textbf{Method} &
\textbf{Backbone} &
\textbf{Image Size} &
\textbf{mIoU} &
\rotatebox{90}{\makecell{\clsbox{cbarrier} barrier}} &
\rotatebox{90}{\makecell{\clsbox{cbicycle} bicycle}} &
\rotatebox{90}{\makecell{\clsbox{cbus} bus}} &
\rotatebox{90}{\makecell{\clsbox{ccar} car}} &
\rotatebox{90}{\makecell{\clsbox{cconsveh} construction\\vehicle}} &
\rotatebox{90}{\makecell{\clsbox{cmotor} motorcycle}} &
\rotatebox{90}{\makecell{\clsbox{cped} pedestrian}} &
\rotatebox{90}{\makecell{\clsbox{ctfcone} traffic cone}} &
\rotatebox{90}{\makecell{\clsbox{ctrailer} trailer}} &
\rotatebox{90}{\makecell{\clsbox{ctruck} truck}} &
\rotatebox{90}{\makecell{\clsbox{cdrvsurf} driveable\\surface}} &
\rotatebox{90}{\makecell{\clsbox{cotherflat} other flat}} &
\rotatebox{90}{\makecell{\clsbox{csidewalk} sidewalk}} &
\rotatebox{90}{\makecell{\clsbox{cterrain} terrain}} &
\rotatebox{90}{\makecell{\clsbox{cmannmade} manmade}} &
\rotatebox{90}{\makecell{\clsbox{cveg} vegetation}}


\\
\midrule
Panoptic-FlashOcc-tiny (1f) & ResNet-50 & 704\,$\times$\,256 & 29.1 & 41.0 & 22.2 & 39.7 & 42.6 & 20.5 & 24.0 & 23.7 & 24.6 & 25.6 & 30.6 & 58.0 & 32.1 & 33.8 & 31.0 & 17.7 & 17.7  \\
Panoptic-FlashOcc (1f) & ResNet-50 & 704\,$\times$\,256 & 29.4 & 42.1 & 22.8 & 40.1 & 42.9 & 20.7 & 24.6 & 23.7 & 24.0 & 25.5 & 30.9 & 58.6 & 32.0 & 34.3 & 31.1 & 18.3 & 17.8  \\
Panoptic-FlashOcc (2f) & ResNet-50 & 704\,$\times$\,256 & 30.3 & 43.9 & 24.4 & 41.9 & 45.2 & 18.7 & 25.6 & 25.7 & 25.9 & 25.3 & 31.8 & 59.0 & 31.5 & 34.7 & 31.5 & 19.9 & 19.3  \\
Panoptic-FlashOcc (8f) & ResNet-50 & 704\,$\times$\,256 & 31.6 & 45.9 & 24.7 & 41.8 & 46.1 & 21.0 & 26.8 & 26.8 & 29.7 & 24.7 & 32.8 & 60.4 & 32.9 & 36.5 & 33.2 & 21.3 & 20.9  \\
HyGE-Occ-tiny (1f)  & ResNet-50 & 704\,$\times$\,256 & 29.4 & 42.0 & 22.3 & 40.8 & 43.9 & 21.9 & 23.8 & 23.7 & 24.4 & 23.5 & 31.5 & 57.0 & 31.4 & 34.2 & 30.4 & 20.1 & 18.5 \\
HyGE-Occ (1f)  & ResNet-50 & 704\,$\times$\,256 & 29.6 & 43.2 & 22.5 & 41.6 & 44.5 & 20.0 & 23.4 & 23.9 & 23.8 & 24.4 & 30.9 & 57.7 & 31.5 & 34.8 & 31.3 & 20.4 & 18.9 \\
HyGE-Occ (2f)  & ResNet-50 & 704\,$\times$\,256 & 30.4 & 43.0 & 23.4 & 42.4 & 45.5 & 21.3 & 25.1 & 25.3 & 25.4 & 25.4 & 32.3 & 58.1 & 30.9 & 34.8 & 31.3 & 20.9 & 19.8 \\
HyGE-Occ (8f)  & ResNet-50 & 704\,$\times$\,256 & 32.0 & 46.7 & 26.6 & 42.8 & 47.2 & 21.6 & 26.6 & 27.0 & 29.9 & 23.6 & 33.5 & 59.6 & 31.3 & 36.6 & 32.7 & 23.9 & 22.4 \\
ALOcc-2D-mini (16f)  & ResNet-50 & 704\,$\times$\,256 & 33.4 & 46.2 & 26.1 & 40.5 & 46.9 & 22.6 & 26.6 & 26.6 & 29.7 & 25.9 & 35.5 & 59.8 & 41.3 & 42.1 & 36.1 & 24.7 & 23.7 \\
ALOcc-2D (16f)  & ResNet-50 & 704\,$\times$\,256 & 37.4 & 48.9 & 30.5 & 42.7 & 49.7 & 27.1 & 29.9 & 29.0 & 35.2 & 32.0 & 38.3 & 63.0 & 44.6 & 45.1 & 39.2 & 34.6 & 30.3 \\
ALOcc-2D-mini (16f) + HyGE-Occ & ResNet-50 & 704\,$\times$\,256 & 33.9 & 45.1 & 27.5 & 42.3 & 47.8 & 24.1 & 28.4 & 26.7 & 30.4 & 27.6 & 35.9 & 58.5 & 41.5 & 42.0 & 35.5 & 25.9 & 23.6 \\
ALOcc-2D (16f) + HyGE-Occ & ResNet-50 & 704\,$\times$\,256 & 37.1 & 50.0 & 30.7 & 42.2 & 50.1 & 25.2 & 32.0 & 29.6 & 36.0 & 30.2 & 38.3 & 61.3 & 42.0 & 43.9 & 38.2 & 36.0 & 30.9 \\

\bottomrule
\end{tabular}
}
\vspace{-1em}
\caption{Per-class IoU score on Occ3D-nuScenes.}
\vspace{-2em}
\label{tab:per_miou}
\end{table*}


\begin{table*}[t]
\centering
\scriptsize
\setlength{\textwidth}{2.8pt}

\newcommand{\clsbox}[1]{\colorbox{#1}{\rule{0pt}{1pt}\rule{1pt}{0pt}}}

\definecolor{cbarrier}{HTML}{FF7832}
\definecolor{cbicycle}{HTML}{FFC0CB}
\definecolor{cbus}{HTML}{FFFF00}
\definecolor{ccar}{HTML}{0096F5}
\definecolor{cconsveh}{HTML}{00FFFF}
\definecolor{cmotor}{HTML}{C8B400}
\definecolor{cped}{HTML}{FF0000}
\definecolor{ctfcone}{HTML}{FFF096}
\definecolor{ctrailer}{HTML}{873C00}
\definecolor{ctruck}{HTML}{A020F0}
\definecolor{cdrvsurf}{HTML}{FF00FF}
\definecolor{cotherflat}{HTML}{AF004B}
\definecolor{csidewalk}{HTML}{4B004B}
\definecolor{cterrain}{HTML}{96F050}
\definecolor{cmannmade}{HTML}{E6E6FA}
\definecolor{cveg}{HTML}{00AF00}

\resizebox{0.98\linewidth}{!}{
\begin{tabular}{l|c|c|c|c|
                *{16}{>{\centering\arraybackslash}p{1.1em}}}
\toprule
\textbf{Method} &
\textbf{Backbone} &
\textbf{Image Size} &
\textbf{RayIoU} &
\textbf{@$d$}&
\rotatebox{90}{\makecell{\clsbox{cbarrier} barrier}} &
\rotatebox{90}{\makecell{\clsbox{cbicycle} bicycle}} &
\rotatebox{90}{\makecell{\clsbox{cbus} bus}} &
\rotatebox{90}{\makecell{\clsbox{ccar} car}} &
\rotatebox{90}{\makecell{\clsbox{cconsveh} construction\\vehicle}} &
\rotatebox{90}{\makecell{\clsbox{cmotor} motorcycle}} &
\rotatebox{90}{\makecell{\clsbox{cped} pedestrian}} &
\rotatebox{90}{\makecell{\clsbox{ctfcone} traffic cone}} &
\rotatebox{90}{\makecell{\clsbox{ctrailer} trailer}} &
\rotatebox{90}{\makecell{\clsbox{ctruck} truck}} &
\rotatebox{90}{\makecell{\clsbox{cdrvsurf} driveable\\surface}} &
\rotatebox{90}{\makecell{\clsbox{cotherflat} other flat}} &
\rotatebox{90}{\makecell{\clsbox{csidewalk} sidewalk}} &
\rotatebox{90}{\makecell{\clsbox{cterrain} terrain}} &
\rotatebox{90}{\makecell{\clsbox{cmannmade} manmade}} &
\rotatebox{90}{\makecell{\clsbox{cveg} vegetation}}


\\
\midrule
BEVFormer & ResNet-101 & 1600\,$\times$\,900 & 33.7 & -- & 42.2 & 18.2 & 55.2 & 57.1 & 22.7 & 21.3 & 31.0 & 27.1 & 30.7 & 49.4 & 58.4 & 30.4 & 29.4 & 31.7 & 36.3 & 26.5 \\
BEVFormer${\dagger}$ & ResNet-101 & 1600\,$\times$\,900 & 32.4 & -- & 44.8 & 24.0 & 55.2 & 56.7 & 21.0 & 29.8 & 33.5 & 26.8 & 27.9 & 49.5 & 45.8 & 18.7 & 22.4 & 18.5 & 39.1 & 29.8 \\
\midrule
FB-Occ & ResNet-50 & 704\,$\times$\,256 & 35.6 & -- & 44.8 & 25.6 & 55.6 & 51.7 & 22.6 & 27.2 & 34.3 & 30.3 & 23.7 & 44.1 & 65.5 & 33.3 & 31.4 & 32.5 & 39.6 & 33.3 \\
FB-Occ${\dagger}$ & ResNet-50 & 704\,$\times$\,256 & 33.5 & -- & 44.9 & 26.2 & 59.7 & 55.1 & 27.9 & 29.1 & 34.3 & 29.6 & 29.1 & 50.5 & 44.4 & 22.4 & 21.5 & 19.5 & 39.3 & 31.1 \\
\midrule
\multirow{3}{*}{Panoptic-FlashOcc-tiny (1f)} & \multirow{3}{*}{ResNet-50} & \multirow{3}{*}{704\,$\times$\,256} & \multirow{3}{*}{34.8} & 29.1 & 37.4 & 20.8 & 49.8 & 48.9 & 17.1 & 19.0 & 29.5 & 29.0 & 20.7 & 41.1 & 53.1 & 28.6 & 23.4 & 22.0 & 26.7 & 17.4  \\
& & & &
35.7 & 42.9 & 24.2 & 60.3 & 56.8 & 24.7 & 27.7 & 34.4 & 32.4 & 29.2 & 50.7 & 61.4 & 32.5 & 28.0 & 29.0 & 34.3 & 27.2 \\
& & & &
39.7 & 45.2 & 24.8 & 65.9 & 59.8 & 27.9 & 29.8 & 36.1 & 33.2 & 36.8 & 55.1 & 70.4 & 35.7 & 32.8 & 35.6 & 39.2 & 35.8 \\
\midrule
\multirow{3}{*}{Panoptic-FlashOcc (1f)} & \multirow{3}{*}{ResNet-50} & \multirow{3}{*}{704\,$\times$\,256} & \multirow{3}{*}{35.2} & 29.4 & 38.7 & 21.8 & 51.4 & 48.7 & 17.6 & 20.3 & 30.1 & 28.0 & 22.7 & 39.5 & 53.4 & 28.9 & 23.4 & 22.2 & 28.0 & 17.6  \\
& & & &
36.0 & 44.2 & 25.7 & 61.3 & 56.4 & 25.4 & 29.2 & 34.9 & 31.3 & 31.3 & 49.3 & 61.8 & 32.6 & 28.0 & 29.1 & 35.1 & 27.3 \\
& & & &
40.1 & 46.5 & 26.5 & 66.9 & 59.2 & 28.8 & 31.0 & 36.6 & 32.1 & 39.0 & 53.7 & 70.8 & 35.6 & 32.9 & 35.6 & 40.1 & 35.9 \\
\midrule
\multirow{3}{*}{Panoptic-FlashOcc (2f)} & \multirow{3}{*}{ResNet-50} & \multirow{3}{*}{704\,$\times$\,256} & \multirow{3}{*}{36.8} & 31.2 & 41.1 & 25.2 & 54.1 & 52.0 & 16.4 & 21.2 & 32.6 & 31.2 & 22.0 & 43.0 & 55.2 & 29.3 & 24.2 & 23.6 & 30.3 & 19.3  \\
& & & &
37.6 & 46.0 & 28.6 & 64.6 & 59.4 & 23.5 & 30.5 & 37.3 & 34.1 & 29.1 & 52.0 & 63.3 & 33.0 & 29.1 & 30.5 & 37.8 & 29.4 \\
& & & &
41.5 & 48.0 & 29.3 & 69.8 & 62.1 & 26.4 & 32.1 & 38.9 & 34.8 & 37.2 & 56.5 & 72.0 & 36.1 & 34.0 & 36.9 & 42.9 & 38.1 \\
\midrule
\multirow{3}{*}{Panoptic-FlashOcc (8f)} & \multirow{3}{*}{ResNet-50} & \multirow{3}{*}{704\,$\times$\,256} & \multirow{3}{*}{38.5} & 32.8 & 43.9 & 25.8 & 54.0 & 53.1 & 18.0 & 24.7 & 34.7 & 34.6 & 20.9 & 45.2 & 56.2 & 29.0 & 26.1 & 26.0 & 34.5 & 22.9  \\
& & & &
39.3 & 48.0 & 28.6 & 64.9 & 60.3 & 25.2 & 32.8 & 39.3 & 37.1 & 29.2 & 54.4 & 64.6 & 32.8 & 31.3 & 33.0 & 42.1 & 33.7 \\
& & & &
43.4 & 49.7 & 29.3 & 70.0 & 62.9 & 28.2 & 34.3 & 40.9 & 37.8 & 38.4 & 58.7 & 73.4 & 36.3 & 36.3 & 39.4 & 47.1 & 42.3 \\
\midrule
\multirow{3}{*}{HyGE-Occ-tiny (1f)} & \multirow{3}{*}{ResNet-50} & \multirow{3}{*}{704\,$\times$\,256} & \multirow{3}{*}{36.3} & 30.8 & 40.0 & 23.4 & 51.9 & 51.0 & 19.2 & 21.2 & 30.9 & 29.7 & 22.1 & 43.4 & 54.4 & 29.6 & 24.5 & 24.1 & 30.6 & 18.6  \\
& & & &
37.2 & 44.6 & 26.5 & 61.8 & 58.5 & 28.0 & 29.3 & 35.6 & 32.3 & 29.7 & 52.6 & 62.4 & 33.1 & 29.4 & 31.0 & 38.5 & 28.2 \\
& & & &
41.1 & 46.5 & 27.1 & 67.2 & 61.3 & 31.7 & 30.7 & 37.2 & 33.0 & 37.6 & 56.7 & 70.9 & 36.1 & 34.4 & 37.3 & 43.4 & 36.4 \\
\midrule
\multirow{3}{*}{HyGE-Occ (1f)} & \multirow{3}{*}{ResNet-50} & \multirow{3}{*}{704\,$\times$\,256} & \multirow{3}{*}{36.8} & 31.2 & 41.2 & 24.5 & 55.2 & 51.4 & 17.2 & 20.8 & 31.4 & 30.5 & 22.2 & 43.0 & 54.9 & 29.7 & 25.0 & 24.7 & 31.0 & 19.2  \\
& & & &
37.5 & 45.6 & 27.2 & 65.0 & 58.7 & 25.1 & 29.0 & 35.9 & 33.0 & 30.4 & 52.4 & 62.9 & 33.4 & 29.8 & 31.7 & 38.9 & 28.6 \\
& & & &
41.6 & 47.4 & 27.8 & 70.2 & 61.7 & 29.2 & 30.3 & 37.5 & 33.7 & 40.5 & 56.4 & 71.4 & 36.6 & 34.6 & 37.8 & 44.0 & 36.7 \\
\midrule
\multirow{3}{*}{HyGE-Occ (2f)} & \multirow{3}{*}{ResNet-50} & \multirow{3}{*}{704\,$\times$\,256} & \multirow{3}{*}{37.8} & 32.2 & 41.9 & 24.9 & 54.7 & 53.4 & 17.9 & 21.3 & 33.5 & 31.8 & 24.0 & 45.4 & 55.5 & 28.8 & 25.9 & 24.9 & 32.6 & 20.6  \\
& & & &
38.6 & 46.7 & 27.8 & 64.8 & 60.6 & 26.3 & 29.5 & 38.0 & 34.6 & 32.9 & 55.0 & 63.5 & 32.7 & 30.9 & 31.8 & 40.7 & 30.6 \\
& & & &
42.6 & 48.3 & 28.5 & 70.0 & 63.0 & 30.7 & 30.7 & 39.6 & 35.3 & 41.8 & 58.6 & 71.9 & 36.0 & 35.8 & 38.2 & 45.9 & 38.9 \\
\midrule
\multirow{3}{*}{HyGE-Occ (8f)} & \multirow{3}{*}{ResNet-50} & \multirow{3}{*}{704\,$\times$\,256} & \multirow{3}{*}{39.9} & 34.5 & 44.8 & 29.3 & 56.2 & 54.9 & 19.9 & 22.8 & 36.3 & 36.4 & 23.5 & 47.6 & 56.9 & 29.1 & 27.3 & 27.0 & 37.9 & 25.1  \\
& & & &
40.7 & 48.7 & 32.2 & 65.2 & 61.7 & 27.7 & 31.4 & 41.1 & 38.9 & 31.9 & 55.8 & 65.2 & 33.0 & 32.4 & 34.3 & 45.8 & 36.1 \\
& & & &
44.5 & 50.3 & 32.8 & 69.9 & 64.1 & 30.2 & 32.9 & 42.7 & 39.5 & 39.8 & 58.9 & 73.9 & 36.8 & 37.4 & 41.0 & 50.7 & 44.4 \\
\midrule
\multirow{3}{*}{ALOcc-2D-mini (16f)} & \multirow{3}{*}{ResNet-50} & \multirow{3}{*}{704\,$\times$\,256} & \multirow{3}{*}{39.3} & 32.9 & 43.0 & 21.7 & 51.0 & 51.5 & 19.3 & 23.3 & 31.1 & 31.0 & 22.7 & 44.8 & 57.3 & 31.7 & 28.4 & 26.4 & 38.0 & 25.8  \\
& & & &
40.1 & 47.8 & 25.4 & 62.1 & 59.4 & 31.3 & 31.7 & 37.5 & 33.7 & 30.8 & 55.5 & 65.4 & 36.6 & 34.1 & 33.8 & 46.6 & 37.5 \\
& & & &
44.8 & 49.7 & 26.7 & 68.2 & 62.6 & 36.9 & 34.0 & 39.7 & 34.9 & 40.9 & 60.8 & 73.9 & 40.9 & 39.5 & 40.8 & 52.3 & 46.4 \\
\midrule
\multirow{3}{*}{ALOcc-2D (16f)} & \multirow{3}{*}{ResNet-50} & \multirow{3}{*}{704\,$\times$\,256} & \multirow{3}{*}{43.0} & 37.1 & 45.5 & 27.2 & 58.0 & 54.6 & 24.1 & 27.5 & 35.0 & 35.0 & 26.8 & 48.3 & 60.9 & 36.0 & 31.7 & 30.6 & 44.7 & 31.9  \\
& & & &
43.8 & 49.4 & 30.6 & 68.2 & 61.6 & 34.6 & 35.8 & 40.4 & 37.2 & 34.8 & 57.9 & 68.4 & 40.8 & 37.5 & 38.2 & 52.4 & 43.6 \\
& & & &
48.2 & 51.0 & 31.8 & 74.1 & 64.7 & 39.2 & 38.6 & 42.5 & 38.2 & 44.7 & 62.5 & 76.0 & 44.9 & 42.7 & 44.6 & 57.4 & 51.8 \\
\midrule
\multirow{3}{*}{ALOcc-2D-mini (16f) + HyGE-Occ} & \multirow{3}{*}{ResNet-50} & \multirow{3}{*}{704\,$\times$\,256} & \multirow{3}{*}{40.2} & 33.9 & 42.5 & 24.4 & 55.8 & 52.5 & 20.3 & 24.8 & 31.7 & 31.9 & 25.7 & 46.2 & 56.5 & 31.8 & 27.9 & 26.3 & 39.5 & 26.3  \\
& & & &
40.9 & 47.6 & 28.0 & 66.5 & 60.1 & 31.3 & 32.1 & 37.6 & 34.3 & 32.6 & 56.7 & 64.8 & 36.8 & 33.8 & 33.8 & 47.7 & 38.1 \\
& & & &
45.6 & 49.7 & 29.1 & 72.3 & 63.3 & 35.5 & 36.4 & 39.7 & 35.6 & 43.2 & 61.7 & 73.6 & 41.3 & 39.5 & 40.8 & 53.0 & 47.2 \\
\midrule
\multirow{3}{*}{ALOcc-2D (16f) + HyGE-Occ} & \multirow{3}{*}{ResNet-50} & \multirow{3}{*}{704\,$\times$\,256} & \multirow{3}{*}{43.2} & 37.3 & 47.1 & 28.3 & 59.1 & 55.6 & 23.6 & 26.4 & 35.8 & 36.8 & 28.0 & 50.3 & 59.2 & 32.5 & 31.2 & 30.1 & 45.4 & 32.9  \\
& & & &
44.0 & 50.5 & 32.5 & 69.3 & 62.7 & 33.7 & 34.8 & 40.7 & 39.1 & 35.8 & 59.2 & 67.2 & 37.5 & 37.0 & 37.7 & 53.5 & 44.4 \\
& & & &
48.3 & 52.3 & 33.4 & 75.0 & 65.4 & 37.5 & 37.6 & 42.5 & 39.9 & 45.7 & 63.1 & 75.4 & 42.0 & 42.3 & 44.3 & 58.6 & 52.7 \\

\bottomrule
\end{tabular}
}
\vspace{-1em}
\caption{Per-class RayIoU score on Occ3D-nuScenes.}
\vspace{-2em}
\label{tab:per_rayiou}
\end{table*}


\begin{table*}[t]
\centering
\scriptsize
\setlength{\textwidth}{2.8pt}

\newcommand{\clsbox}[1]{\colorbox{#1}{\rule{0pt}{1pt}\rule{1pt}{0pt}}}

\definecolor{cbarrier}{HTML}{FF7832}
\definecolor{cbicycle}{HTML}{FFC0CB}
\definecolor{cbus}{HTML}{FFFF00}
\definecolor{ccar}{HTML}{0096F5}
\definecolor{cconsveh}{HTML}{00FFFF}
\definecolor{cmotor}{HTML}{C8B400}
\definecolor{cped}{HTML}{FF0000}
\definecolor{ctfcone}{HTML}{FFF096}
\definecolor{ctrailer}{HTML}{873C00}
\definecolor{ctruck}{HTML}{A020F0}
\definecolor{cdrvsurf}{HTML}{FF00FF}
\definecolor{cotherflat}{HTML}{AF004B}
\definecolor{csidewalk}{HTML}{4B004B}
\definecolor{cterrain}{HTML}{96F050}
\definecolor{cmannmade}{HTML}{E6E6FA}
\definecolor{cveg}{HTML}{00AF00}

\resizebox{0.98\linewidth}{!}{
\begin{tabular}{l|c|c|c|c|
                *{16}{>{\centering\arraybackslash}p{1.1em}}}
\toprule
\textbf{Method} &
\textbf{Backbone} &
\textbf{Image Size} &
\textbf{RayPQ} &
\textbf{@$d$} &
\rotatebox{90}{\makecell{\clsbox{cbarrier} barrier}} &
\rotatebox{90}{\makecell{\clsbox{cbicycle} bicycle}} &
\rotatebox{90}{\makecell{\clsbox{cbus} bus}} &
\rotatebox{90}{\makecell{\clsbox{ccar} car}} &
\rotatebox{90}{\makecell{\clsbox{cconsveh} construction\\vehicle}} &
\rotatebox{90}{\makecell{\clsbox{cmotor} motorcycle}} &
\rotatebox{90}{\makecell{\clsbox{cped} pedestrian}} &
\rotatebox{90}{\makecell{\clsbox{ctfcone} traffic cone}} &
\rotatebox{90}{\makecell{\clsbox{ctrailer} trailer}} &
\rotatebox{90}{\makecell{\clsbox{ctruck} truck}} &
\rotatebox{90}{\makecell{\clsbox{cdrvsurf} driveable\\surface}} &
\rotatebox{90}{\makecell{\clsbox{cotherflat} other flat}} &
\rotatebox{90}{\makecell{\clsbox{csidewalk} sidewalk}} &
\rotatebox{90}{\makecell{\clsbox{cterrain} terrain}} &
\rotatebox{90}{\makecell{\clsbox{cmannmade} manmade}} &
\rotatebox{90}{\makecell{\clsbox{cveg} vegetation}}


\\
\midrule
\multirow{3}{*}{Panoptic-FlashOcc-tiny (1f)} & \multirow{3}{*}{ResNet-50} & \multirow{3}{*}{704\,$\times$\,256} & \multirow{3}{*}{12.9} & 8.8 & 10.7 & 6.9 & 24.4 & 23.8 & 1.8 & 6.1 & 1.6 & 3.0 & 2.9 & 15.1 & 40.7 & 5.4 & 0.9 & 0.6 & 4.4 & 0.1  \\
& & & &
13.4 & 16.9 & 8.6 & 35.0 & 32.6 & 8.1 & 10.5 & 2.2 & 4.9 & 4.7 & 24.0 & 53.1 & 7.8 & 3.0 & 2.2 & 9.1 & 2.1 \\
& & & &
16.5 & 20.4 & 8.8 & 40.8 & 35.2 & 10.5 & 11.7 & 2.3 & 5.2 & 5.6 & 28.6 & 66.2 & 9.8 & 6.1 & 4.5 & 12.8 & 9.1 \\
\midrule
\multirow{3}{*}{Panoptic-FlashOcc (1f)} & \multirow{3}{*}{ResNet-50} & \multirow{3}{*}{704\,$\times$\,256} & \multirow{3}{*}{13.2} & 9.2 & 12.5 & 5.1 & 27.5 & 24.2 & 1.6 & 7.1 & 1.7 & 3.2 & 3.5 & 14.5 & 41.0 & 6.2 & 0.8 & 1.0 & 5.4 & 0.3  \\
& & & &
13.5 & 18.2 & 7.2 & 36.6 & 33.2 & 5.8 & 12.4 & 2.2 & 4.0 & 5.5 & 23.2 & 53.7 & 8.7 & 3.0 & 2.6 & 9.1 & 2.2 \\
& & & &
16.8 & 21.8 & 7.6 & 42.2 & 35.6 & 9.2 & 13.7 & 2.3 & 4.4 & 6.3 & 28.2 & 66.5 & 10.9 & 6.4 & 4.7 & 13.4 & 9.2 \\
\midrule
\multirow{3}{*}{Panoptic-FlashOcc (2f)} & \multirow{3}{*}{ResNet-50} & \multirow{3}{*}{704\,$\times$\,256} & \multirow{3}{*}{14.5} & 10.6 & 15.5 & 8.3 & 29.9 & 27.7 & 1.1 & 9.8 & 2.1 & 5.2 & 4.3 & 15.8 & 44.0 & 6.5 & 1.2 & 0.9 & 6.0 & 0.1  \\
& & & &
15.0 & 21.1 & 9.7 & 39.1 & 36.0 & 6.2 & 14.9 & 2.6 & 6.9 & 6.2 & 24.8 & 55.9 & 8.9 & 2.9 & 2.8 & 10.8 & 2.9 \\
& & & &
18.0 & 23.5 & 10.2 & 44.2 & 38.4 & 7.7 & 16.6 & 2.7 & 7.1 & 7.1 & 29.3 & 68.0 & 10.7 & 6.0 & 5.3 & 15.3 & 11.1 \\
\midrule
\multirow{3}{*}{Panoptic-FlashOcc (8f)} & \multirow{3}{*}{ResNet-50} & \multirow{3}{*}{704\,$\times$\,256} & \multirow{3}{*}{16.0} & 11.9 & 18.4 & 8.8 & 31.1 & 30.0 & 3.2 & 11.4 & 2.5 & 7.1 & 4.9 & 18.2 & 45.7 & 6.2 & 1.8 & 1.7 & 7.7 & 0.2  \\
& & & &
16.3 & 23.2 & 10.3 & 40.6 & 38.0 & 5.7 & 15.6 & 3.0 & 8.1 & 7.7 & 27.4 & 57.4 & 8.6 & 4.2 & 3.9 & 14.4 & 6.1 \\
& & & &
19.7 & 25.3 & 10.8 & 45.8 & 40.3 & 8.1 & 16.9 & 3.1 & 8.5 & 8.8 & 31.4 & 70.2 & 10.6 & 9.1 & 7.4 & 19.4 & 16.2 \\
\midrule
\multirow{3}{*}{HyGE-Occ-tiny (1f)} & \multirow{3}{*}{ResNet-50} & \multirow{3}{*}{704\,$\times$\,256} & \multirow{3}{*}{14.5} & 10.5 & 14.6 & 9.1 & 28.9 & 26.7 & 2.8 & 9.5 & 1.9 & 4.4 & 3.6 & 16.8 & 42.6 & 7.1 & 1.2 & 1.5 & 6.2 & 0.3  \\
& & & &
14.9 & 20.3 & 10.4 & 38.5 & 35.0 & 8.2 & 13.7 & 2.4 & 5.9 & 5.3 & 25.6 & 54.5 & 8.9 & 3.4 & 3.6 & 11.4 & 3.5 \\
& & & &
18.1 & 23.2 & 10.6 & 44.0 & 37.4 & 10.8 & 15.1 & 2.5 & 6.4 & 6.4 & 30.2 & 66.5 & 10.9 & 7.9 & 6.1 & 16.0 & 10.7 \\
\midrule
\multirow{3}{*}{HyGE-Occ (1f)} & \multirow{3}{*}{ResNet-50} & \multirow{3}{*}{704\,$\times$\,256} & \multirow{3}{*}{15.1} & 11.1 & 15.2 & 10.1 & 31.3 & 27.2 & 2.8 & 10.6 & 2.1 & 4.8 & 4.3 & 16.6 & 43.3 & 7.6 & 1.4 & 1.9 & 6.4 & 0.4  \\
& & & &
15.5 & 21.5 & 11.4 & 40.9 & 35.4 & 6.6 & 16.0 & 2.6 & 6.4 & 6.4 & 25.4 & 55.1 & 9.5 & 3.6 & 3.8 & 11.9 & 3.7 \\
& & & &
18.9 & 24.2 & 11.6 & 46.1 & 38.1 & 11.1 & 17.6 & 2.7 & 6.6 & 7.7 & 30.2 & 67.5 & 11.6 & 8.1 & 6.6 & 16.7 & 11.4 \\
\midrule
\multirow{3}{*}{HyGE-Occ (2f)} & \multirow{3}{*}{ResNet-50} & \multirow{3}{*}{704\,$\times$\,256} & \multirow{3}{*}{15.8} & 11.7 & 16.8 & 10.7 & 31.0 & 29.7 & 2.4 & 12.2 & 2.4 & 5.4 & 4.2 & 18.7 & 44.4 & 6.9 & 1.9 & 1.5 & 7.1 & 0.5  \\
& & & &
16.1 & 22.0 & 12.2 & 39.8 & 37.6 & 6.4 & 16.9 & 2.8 & 7.1 & 6.2 & 27.1 & 55.9 & 9.9 & 5.1 & 3.7 & 13.7 & 5.1 \\
& & & &
19.5 & 24.1 & 12.6 & 45.9 & 39.8 & 10.8 & 18.4 & 2.9 & 7.6 & 7.6 & 31.6 & 68.0 & 11.6 & 9.2 & 6.6 & 18.6 & 13.3 \\
\midrule
\multirow{3}{*}{HyGE-Occ (8f)} & \multirow{3}{*}{ResNet-50} & \multirow{3}{*}{704\,$\times$\,256} & \multirow{3}{*}{17.5} & 13.1 & 20.9 & 11.3 & 33.1 & 32.3 & 2.4 & 14.3 & 2.5 & 8.4 & 4.3 & 20.9 & 46.8 & 7.0 & 2.5 & 2.8 & 11.0 & 0.8  \\
& & & &
17.9 & 26.4 & 12.8 & 42.3 & 39.8 & 6.8 & 18.4 & 3.0 & 9.9 & 6.3 & 29.2 & 58.9 & 9.2 & 5.3 & 5.5 & 18.6 & 9.5 \\
& & & &
21.4 & 28.9 & 13.2 & 46.9 & 42.0 & 8.6 & 19.5 & 3.2 & 10.4 & 8.0 & 32.9 & 70.8 & 11.7 & 10.4 & 8.8 & 24.5 & 20.6 \\

\bottomrule
\end{tabular}
}
\vspace{-1em}
\caption{Per-class RayPQ score on Occ3D-nuScenes.}
\vspace{-2em}
\label{tab:per_raypq}
\end{table*}
\clearpage
\clearpage

{
    \small
    \bibliographystyle{ieeenat_fullname}
    \bibliography{supp_main,main}
}
